\documentclass[10pt,twocolumn,letterpaper]{article}

\usepackage[pagenumbers]{iccv} 
\usepackage{graphicx}
\usepackage{bbm}
\usepackage{float}
\usepackage{bbding}
\usepackage{colortbl}
\usepackage{algorithm}
\usepackage{listings}
\usepackage{multirow}
\usepackage{makecell}
\usepackage[resetlabels,labeled]{multibib}
\newcites{S}{References for the Supplementary Material}

\definecolor{iccvblue}{rgb}{0.21,0.49,0.74}
\usepackage[pagebackref,breaklinks,colorlinks,allcolors=iccvblue]{hyperref}

\def\paperID{1091} 
\def\confName{ICCV}
\def\confYear{2025}

\title{Mamba-3D as Masked Autoencoders for Accurate and Data-Efficient \linebreak Analysis of Medical Ultrasound Videos}

\author{Jiaheng Zhou$^{1,2}$ \quad Yanfeng Zhou$^{1,2}$ \quad Wei Fang $^{3,4}$ \quad Yuxing Tang $^{3}$ \quad Le Lu $^{3}$ \quad Ge Yang$^{1,2}$ \vspace{0.3em} \\
{\normalsize $^1$
\normalsize Institute of Automation, Chinese Academy of Sciences} \\
{\normalsize $^2$
School of Artificial Intelligence, University of Chinese Academy of Sciences} \\
{\normalsize $^3$
DAMO Academy, Alibaba Group}
{\normalsize $^4$
Hupan Laboratory, 310023, Hangzhou, China}
}

\begin{document}
\maketitle

\begin{abstract}
Ultrasound videos are an important form of clinical imaging data, and deep learning-based automated analysis can improve diagnostic accuracy and clinical efficiency. However, the scarcity of labeled data and the inherent challenges of video analysis have impeded the advancement of related methods.
In this work, we introduce \textbf{E-ViM³}, a data-efficient \textbf{Vi}sion \textbf{M}amba network that preserves the \textbf{3D} structure of video data, enhancing long-range dependencies and inductive biases to better model space-time correlations. With our design of \textbf{E}nclosure Global Tokens (EGT), the model captures and aggregates global features more effectively than competing methods.
To further improve data efficiency, we employ masked video modeling for self-supervised pre-training, with the proposed Spatial-Temporal Chained (STC) masking strategy designed to adapt to various video scenarios.
Experiments demonstrate that E-ViM³ performs as the state-of-the-art in two high-level semantic analysis tasks across four datasets of varying sizes: EchoNet-Dynamic, CAMUS, MICCAI-BUV, and WHBUS. Furthermore, our model achieves competitive performance with limited labels, highlighting its potential impact on real-world clinical applications. Our code is available at {\small \url{https://github.com/HenryZhou19/E-ViM3}}.
\end{abstract}

\begin{figure}[ht]
  \centering
  \includegraphics[width=0.95\linewidth]{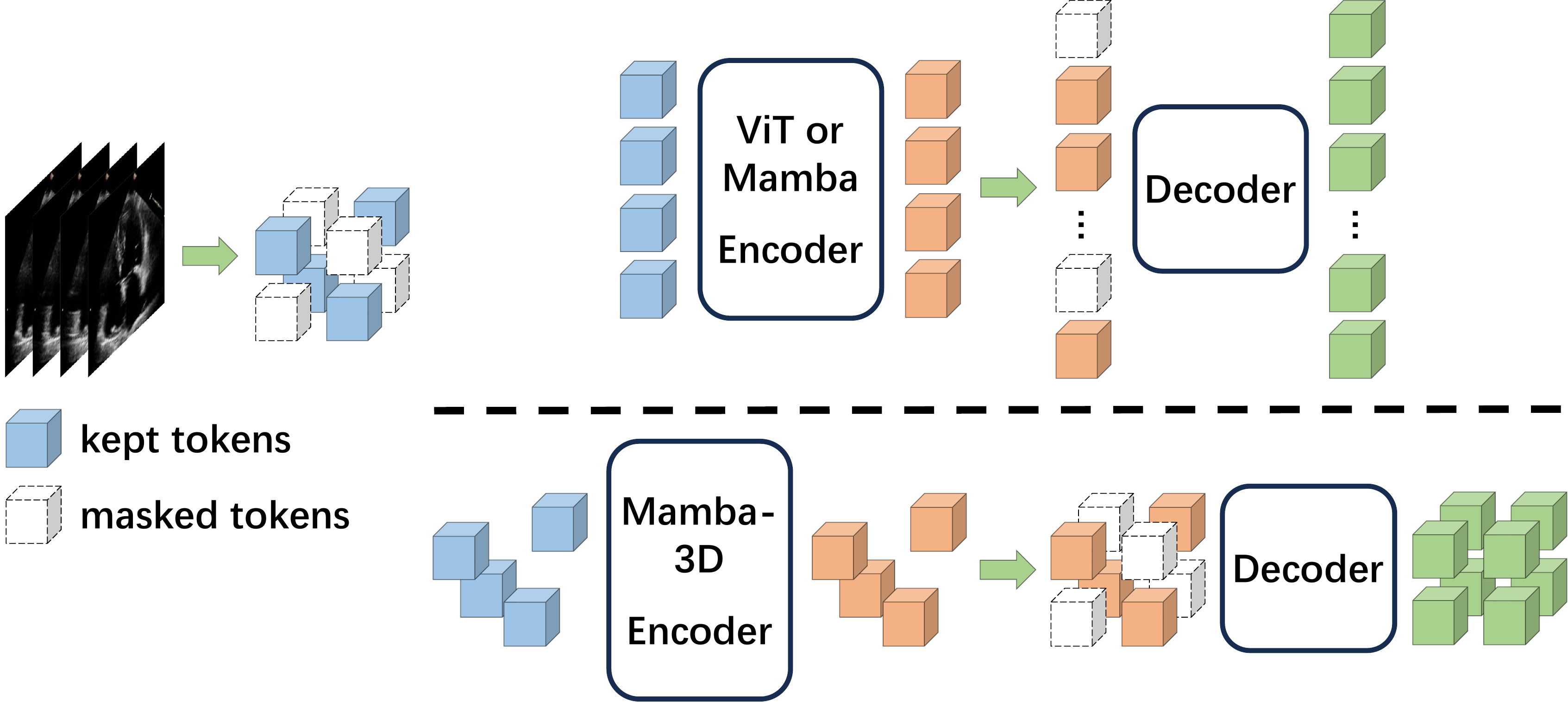}
  \caption{\textbf{Mamba-3D as masked autoencoders.} By preserving the structure of video data, Mamba-3D provides a stronger inductive bias for masked video modeling than ViT~\cite{Dosovitskiy1, Tong1} or vanilla Mamba for visual data~\cite{Zhu1, Liu3, Li1}, both of which operate on flattened 1D sequences and rely heavily on positional encodings. This enables effective self-supervised learning with limited data.}
  \vspace{-0.3cm}
  \label{fig:teaser}
\end{figure}

\begin{figure*}[t]
\centering
\includegraphics[width=\textwidth]{figs/pipeline_bold_final.pdf}
\caption{\textbf{The pipeline of self-supervised pre-training and fine-tuning of the proposed E-ViM³ model. The upper section represents the pre-training phase; the lower section represents the fine-tuning phase.}
Initially, the video is embedded as 3D patches but not directly flattened into a 1D sequence. During pre-training, the proposed Spatial-Temporal Chained masking is applied, and masked tokens are removed for efficiency. The 3D-structured tokens, along with the proposed Enclosure Global Tokens, are inputted into the Mamba-3D encoder. The pre-training task is to restore the masked patches using a decoder, while downstream tasks leverage the features extracted from the global tokens. Further details can be found in~\cref{sec:methods}. A complete architecture diagram and the experiments on left ventricular segmentation are available in the supplementary material.}
\label{fig:pipeline}
\end{figure*}

\section{Introduction}
\label{sec:intro}

In medical imaging, ultrasound video is a widely used modality that requires real-time clinical interpretation, in which missed or incorrect diagnoses often occur due to various human factors, underscoring the importance of AI-assisted analysis~\cite{Shen1}. However, the lack of labeled data has greatly hindered research progress in this field as it often leads to serious overfitting issues in practice. 


Pioneering works by Ouyang~\etal~\cite{Ouyang1, Ouyang2} have established a well-labeled and relatively large dataset of cardiac ultrasound videos, making a significant contribution to related research. However, for many other types of medical ultrasound videos, the lack of high-quality public datasets remains a real challenge.
Therefore, it is both important and practical to develop powerful and data-efficient models for analyzing medical ultrasound videos.

Self-supervised pre-training has been proven to be effective in learning visual features without specific labels, demonstrating feasibility in various scenarios involving images~\cite{He1, Caron1, He2, Bao1} and videos~\cite{Han1, Han2, Wang2}. Most approaches utilize either contrastive learning~\cite{Chen1, Chen2, Pan1} or Transformer-based masked autoencoding~\cite{WEi1, Tong1, Wang1}. However, the reliance of contrastive learning on negative samples makes it unsuitable for medical video applications, as there is a natural dominance of similarities between different samples.
Consequently, this paper focuses on masked video modeling as the pretext task.

Furthermore, Transformer-based models lack inherent inductive biases and heavily rely on large datasets to learn meaningful visual patterns~\cite{Dosovitskiy1, Touvron1}. Additionally, Transformers often struggle with long or high-resolution videos due to the quadratic complexity of their attention mechanism~\cite{Vaswani1, Arnab1, Han3}, and existing improvements are made at the expense of representation quality and convergence efficiency~\cite{Shen2, Katharopoulos1}. These limitations can lead to inadequate training and increased hardware constraints, ultimately affecting performance.

The recently emerged selective state space model, commonly known as Mamba~\cite{Dao1, Gu1}, offers a promising solution to these challenges. Mamba employs latent states and the selective scanning mechanism to model relationships between input and output sequences, providing a more explicit inductive bias for sequential data than Transformers. In addition, with linear complexity and hardware-efficient implementation, Mamba is well-suited for handling long sequences.
However, Mamba blocks use 1D scanning instead of attention, making the positioning of global tokens particularly important, especially when handling long sequences derived from 3D data. And effectively integrating Mamba's 1D scanning with masked video modeling of 3D data remains an open challenge.

In this work, we present E-ViM³, a Mamba-3D network with Enclosure Global Tokens (EGT) for data-efficient analysis of medical ultrasound videos. By employing the proposed Spatial-Temporal Chained (STC) masked video modeling as an effective pretext task, we pre-train the E-ViM³ model and fine-tune it entirely on high-level semantic downstream tasks. Experimental results demonstrate that our model, combined with this training strategy, surpasses previous state-of-the-art methods in left ventricular ejection fraction (EF) prediction and breast cancer classification across four public datasets: EchoNet-Dynamic, CAMUS, MICCAI-BUV, and WHBUS. Furthermore, our method is label-efficient, as it enables the analysis of medical ultrasound videos even with extremely limited labeled data. 

Our main contributions are as follows:

\begin{itemize}

\item We introduce E-ViM³, a new Mamba-3D network incorporating Enclosure Global Tokens (EGT) that effectively aggregates global features on medical ultrasound videos, while maintaining the ability to leverage correlations in both time and space.

\item By caching index transformations between sequences after removing masked tokens, we provide an efficient way to pre-train Mamba-3D blocks as a masked autoencoder.

\item We propose Spatial-Temporal Chained (STC) masking, a customized and easy-to-implement masking strategy that enhances the effectiveness of self-supervised pre-training for different medical ultrasound datasets.

\item Fine-tuning with limited labeled data from EchoNet-Dynamic validates that E-ViM³ is robust under label-scarce conditions. Moreover, results on small datasets such as CAMUS and MICCAI-BUV demonstrate strong generalization and data efficiency over prior methods.

\end{itemize}

\section{Related Works}
\label{sec:related_works}

\subsection{Analysis of Medical Ultrasound Videos}

In echocardiography, ejection fraction (EF) is crucial for diagnosing heart diseases. Studies typically segment the end-diastole (ED) and end-systole (ES) phases for EF calculation~\cite{Wei2, Wei3, Leclerc1, Liu1, Thomas1}, relying on additional label information. Recent efforts have been focusing on developing end-to-end EF regression models for practical clinical applications.

Ouyang \etal~\cite{Ouyang1, Ouyang2} introduced the EchoNet-Dynamic dataset and a baseline method using R(2+1)D~\cite{Tran1} for EF prediction, but their best results relied on an extra segmentation network. Reynaud \etal~\cite{Reynaud1} treated each frame as a single token in the Transformer for arbitrary lengths. 
Mokhtari \etal~\cite{Mokhtari2} further experimented with a multi-level Transformer for multiple video aggregation of the same patient, achieving improvements at the cost of generality.
Muhtaseb \etal~\cite{Muhtaseb1} integrated CNNs and Transformers for better performance.
Additionally,~\cite{Kazemi1, Mokhtari1, Maani1} aimed to enhance interpretability but yielded minimal accuracy gains.

Breast cancer classification is another important application of medical ultrasound videos. Similar to echocardiographic studies, most works relied on additional information like tumor locations, limiting practical applicability~\cite{Sun1, Huang1, chen3, Huang2}. Lin \etal~\cite{Lin1} explored self-supervised pre-training on a large private dataset. Zhang \etal~\cite{Zhang1} sampled short video clips and used clip-level attention to better leverage clips with different qualities.

The research community has sought diverse methods for analyzing ultrasound videos. However, overfitting on current datasets severely affects results.
In this work, we construct a novel self-supervised pre-training network to reduce the dependence on labels and improve the performance. 

\subsection{Mamba Structure for Videos}
As Mamba~\cite{Dao1, Gu1} becomes a strong competitor to Transformer~\cite{Vaswani1} in sequence modeling, several foundational Mamba-based works have been developed for videos.

VideoMamba~\cite{Li1} simply replaced the vanilla ViT block~\cite{Dosovitskiy1} with the bi-directional Mamba block~\cite{Zhu1}, employing a spatial-first scanning mechanism to adapt Mamba for 3D video data. This approach demonstrated Mamba's efficiency for high-resolution, long-duration videos.
Concurrently,~\cite{Park1} proposed a dual scanning method with a similar data flow. These are progressive attempts to use Mamba for videos, but flattening the entire video into a 1D vector disrupts the inherent structure of videos.

Mamba-ND~\cite{Li2} improved Mamba's scanning by varying the flattening order of input sequences, enabling more efficient modeling of high-dimensional data. However, this approach complicated the integration of global tokens and thus relied on pooling all tokens for downstream tasks.


In this work, we follow Mamba-ND~\cite{Li2} to preserve the 3D structure of video data, with Enclosure Global Tokens to facilitate better aggregation of video features.

\subsection{Masked Video Modeling for Self-supervised Pre-training}
Masked video modeling (MVM) is an intuitive generative pretext task and has attracted significant attention in self-supervised pre-training of natural videos.

Most recent MVM works primarily utilize Vision Transformer~\cite{Dosovitskiy1} and its variants~\cite{Touvron2, Liu2}. 
As BEiT~\cite{Bao1} for images, BEVT~\cite{Wang2} aimed to tokenize and reconstruct masked video patches with discrete embeddings from a pre-trained image VQ-VAE~\cite{Ramesh1}. 
MAE-ST~\cite{Feichtenhofer1} and VideoMAE~\cite{Tong1, Wang1} are adaptations of MAE~\cite{He2} for videos, with slight differences in masking strategies. Distinct from BEVT, they are pre-trained by directly predicting the masked pixels, enhancing efficiency and versatility by removing the tokenizer. AdaMAE~\cite{Bandara1} used the policy gradient algorithm to mask more tokens with poor spatiotemporal information. FocusMAE~\cite{Basu1} achieved better gallbladder cancer detection performance through a higher masking priority in the labeled malignant regions. VideoMamba~\cite{Li1} also provided a self-supervised pre-training approach inspired by UMT~\cite{Li3}, which was used for multi-modal video tasks.

In this work, we enhance MVM by utilizing Mamba-3D for medical ultrasound video datasets, which are typically too small for effective training with standard Transformers.

\section{E-ViM³}
\label{sec:methods}

In this section, we first introduce the overall architecture of E-ViM³ with Enclosure Global Tokens (EGT) detailed in~\cref{methods_sub1}.
Next, we discuss the Spatial-Temporal Chained (STC) masked pre-training for Mamba-3D in~\cref{methods_sub2}.
Finally, we specify some main configurations for the two-stage training procedure in~\cref{methods_sub3}. Preliminaries on selective state space models and other detailed configurations are provided in the supplementary material.

\subsection{\label{methods_sub1}Overall Architecture}
The main pipeline of our method is illustrated in~\cref{fig:pipeline}. Similar to other approaches that rely on video patchification~\cite{Bertasius1, Li1, Arnab1, Liu2}, we first embed the video into tokens, incorporating global tokens and positional encoding. This is followed by several cascaded blocks as the encoder, specifically Mamba-3D blocks we employed. The main distinction between pre-training and fine-tuning lies in the choice of networks for the pretext task and the actual downstream tasks, as well as whether masking is applied to the input.

\paragraph{Video embedding}
Consider an arbitrary standard video clip $\mathbf{V}\in\mathbb{R}^{\mathrm{C\times L\times H\times W}}$ as the input, where $\mathrm{C}$ denotes the original number of channels; $\mathrm{L}$, $\mathrm{H}$, and $\mathrm{W}$ represent the number of frames and the spatial resolution, respectively. We first employ a simple 3D embedding layer as the tokenizer to obtain the embedded video $\mathbf{V}_{\text{emb}}\in\mathbb{R}^{\mathrm{D\times L_p\times H_p\times W_p}}$, where $\mathrm{D}$ is the embedding dimension, and the subscript ${}_\mathrm{P}$ stands for \textit{patch}. The 3D embedding layer $f: \mathrm{D^{\prime} \rightarrow D}$ ($\mathrm{D^{\prime}=C\times p_t\times p_s\times p_s}$) is controlled by hyper-parameters $\rm p_t$ and $\rm p_s$, which are the patch sizes in the temporal and spatial dimensions, respectively. Consequently, we have $\mathrm{L_p}=\mathrm{L}/\rm p_t$, $\mathrm{H_p}=\mathrm{H}/\rm p_s$, $\mathrm{W_p}=\mathrm{W}/\rm p_s$. We also keep the original video in patch form as $\mathbf{V}_{\text{patch}}\in\mathbb{R}^{\mathrm{D^{\prime}\times L_p\times H_p\times W_p}}$ for the pretext task.

\begin{figure}[t]
  \centering
  \includegraphics[width=\linewidth]{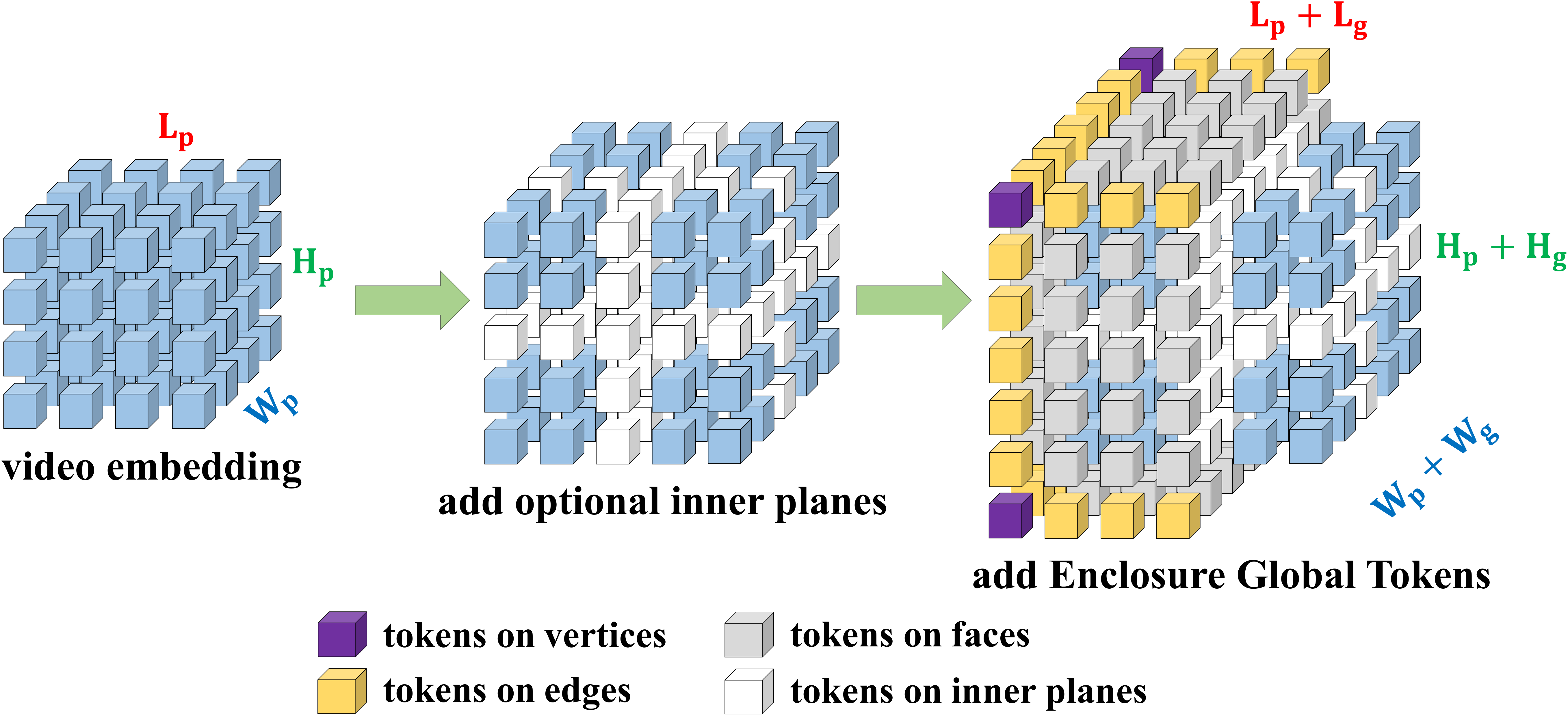}
  \caption{\textbf{An illustration of adding Enclosure Global Tokens to the original video embedding.} The optional inner planes are also taken into account. In this case, we use $\mathrm{L_g=H_g=W_g=3}$.}
  \label{fig:EGTs}
\end{figure}
\vspace{-8pt}

\paragraph{Enclosure Global Tokens}
Global feature aggregation is critical for tokenized data,
particularly in Mamba which is sequence-sensitive. In contrast to Mamba-ND~\cite{Li2}, which utilizes average pooling instead of global tokens for downstream tasks, we propose the Enclosure Global Tokens (EGT) as a part of our E-ViM³. 

As shown in~\cref{fig:EGTs}, the global tokens are intuitively added in an enclosed shell pattern, optionally with evenly distributed planes inside the $\mathbf{V}_{\text{emb}}$ cube, dividing the entire cube into smaller ones. The proposed EGTs, most of which work as registers and correspond to perpendicular planes in the 3D structure, enable visual tokens to deliver information within the shortest distance in the Mamba-3D architecture. 

The final shape of the input tensor to the backbone (\ie encoder) should be:
\begin{equation}
\begin{aligned}
\mathbf{V}_{\text{token}}\in\mathbb{R}^{\mathrm{D\times (L_p+L_g)\times (H_p+H_g)\times (W_p+W_g)}}
\end{aligned}\label{eq:V_token}
\end{equation}
where $\mathrm{L_g}\ge2,\ \mathrm{H_g}\ge2,\ \mathrm{W_g}\ge2$. To balance the number of parameters and the discrimination of different global tokens, we initialize all global tokens with 26 or 27 different sets of learnable parameters, based on the cube's faces (6), edges (12), vertices (8), and optionally the inner planes (1).

While the sequential order is inherent in Mamba's 1D scanning procedure, positional encoding is still essential for Mamba to better handle the spatial-temporal relationship as each scanning goes along the 3D data. Following~\cite{Arnab1}, we employ learnable positional embeddings, with three independent components to adapt to the three dimensions of video data:  $\mathbf{P}_\mathrm{L}\in\mathbb{R}^{\mathrm{D\times(L_p+L_g)}\times1\times1}$, $\mathbf{P}_\mathrm{H}\in\mathbb{R}^{\mathrm{D\times1\times(H_p+H_g)}\times1}$, and $\mathbf{P}_\mathrm{W}\in\mathbb{R}^{\mathrm{D\times1\times1\times(W_p+W_g)}}$.The input to the encoder is then given by:
\begin{equation}
\begin{aligned}
\mathbf{V}^{(0)}_{\text{in}}&=\mathbf{V}_{\text{token}}+\mathbf{P}_\mathrm{L}+\mathbf{P}_\mathrm{H}+\mathbf{P}_\mathrm{W}
\\
&\in\mathbb{R}^{\mathrm{D\times L_{in}\times H_{in}\times W_{in}}}
\end{aligned}\label{eq:V_in}
\end{equation}
where $\mathrm{L_{in}=L_p+L_g},\ \mathrm{H_{in}=H_p+H_g},\ \mathrm{W_{in}=W_p+W_g}$.

\begin{figure}[t]
  \centering
  \includegraphics[width=\columnwidth]{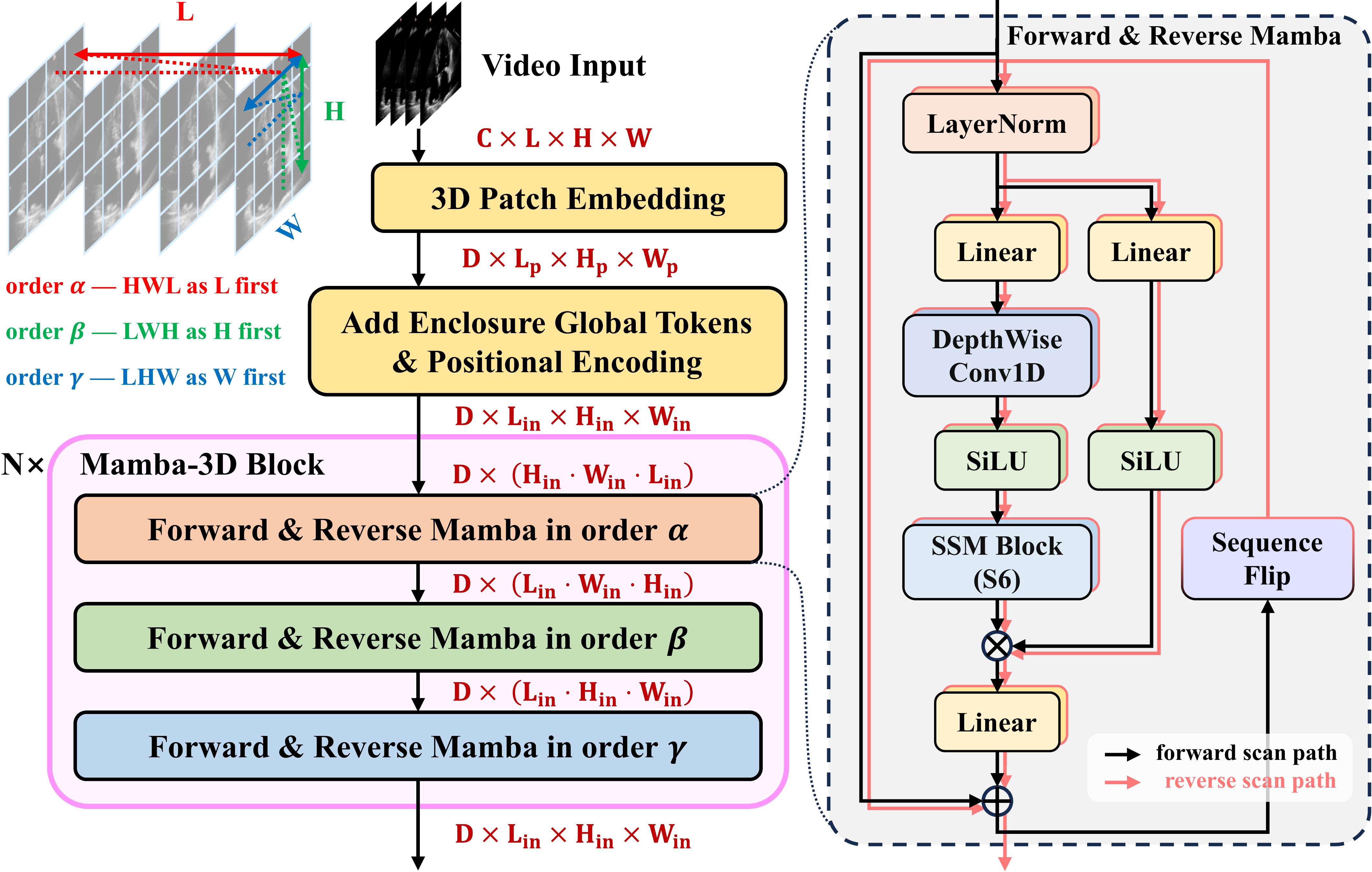}
  \caption{\textbf{The main data flow: from the input video to the encoder's output without token masking.} Please note the changes in data dimensions and their order. Global tokens are omitted on the right for simplicity. Diagrams with more details are available in the supplementary material.}
  \label{fig:mamba-3d}
\end{figure}

\paragraph{Mamba-3D as basic blocks}
All video tokens $\mathbf{V}^{(0)}_{\text{in}}$ are then fed into the backbone, which is designed with an auto-regressive structure consisting of $N$ cascaded blocks, as shown in~\cref{fig:mamba-3d}. Thus, the output of each block $\mathbf{V}^{(n)}_{\mathrm{out}}\ (n=0,1,\dots,N-1)$ maintains the same shape as its input $\mathbf{V}^{(n)}_{\text{in}}$. The final output of the backbone (\ie, $\mathbf{V}^{(N-1)}_{\mathrm{out}}$), or parts of it, will be used for the pretext task or downstream tasks.

The scanning mechanism is critical to models involving Mamba. Unlike existing methods that combine multi-directional scans in parallel into one SSM module~\cite{Zhu1, Li1}, concatenating vanilla Mamba layers in serial with varied directional scans deepens the network without adding too many parameters. Following Mamba-ND~\cite{Li2}, we employ a 6-direction scanning mechanism for 3D data, concatenating independent forward and reverse Mamba on three permutations of 3D axes: “$\rm HWL$”, “$\rm LWH$”, “$\rm LHW$”, where each final axis remains contiguous after flattening. This enhances the exploitation of correlations across all dimensions.

\vspace{-8pt}

\paragraph{Heads for downstream tasks}
We utilize EGTs to aggregate features for high-level semantic tasks. Given the trade-offs in model complexity, we intuitively select the intersection of all extra planes in $\mathbf{V}^{(N-1)}_{\mathrm{out}}$ as key positions, applying a simple concatenation operator across the $\mathrm{L_g\times H_g\times W_g}$ global tokens. Notably, when no inner planes are added, the final features come from the cube's eight vertices. Finally, several fully connected layers form the heads.

\subsection{\label{methods_sub2}Masked Video Modeling for Mamba-3D with Enclosure Global Tokens}

We use Masked Video Modeling (MVM) as the pretext task for self-supervised pre-training, where the decoder is structurally similar to the encoder but much shallower (\ie, the decoder consists $M$ cascaded blocks, with $M\ll N$).

\begin{figure}[t]
  \centering
  \includegraphics[width=\columnwidth]{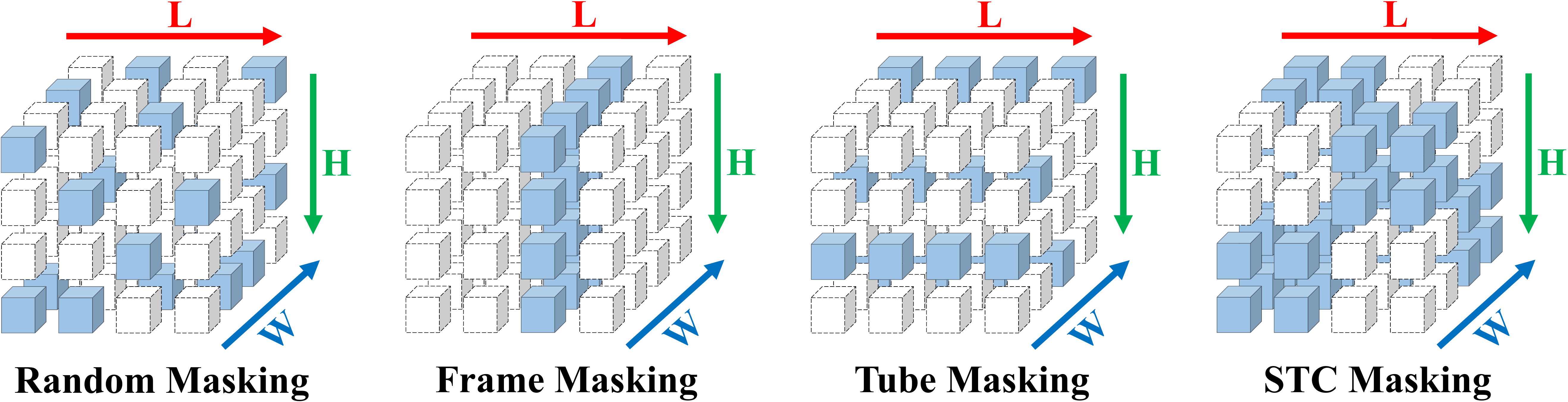}
  \caption{\textbf{Different masking strategies, including the proposed Spatial-Temporal Chained (STC) masking.} Here, we use $\mathrm{\gamma_t=\gamma_s=2}$ as an example for STC. The other three widely used strategies are special cases of STC with specific hyper-parameters: $\mathrm{\gamma_t=\gamma_s=1}$ for Random (Agnostic), $\mathrm{\gamma_t=1, \gamma_s=H=W}$ for Frame, and $\mathrm{\gamma_t=L, \gamma_s=1}$ for Tube.}
  \label{fig:STC}
\end{figure}
\vspace{-8pt}

\paragraph{Spatial-Temporal Chained masking}
Different from the methods given by \cite{Feichtenhofer1, Wang1, Xing1}, which prefer either tube masking or a completely agnostic approach, we propose a more flexible masking strategy tailored for scenarios where the video content has particular variation speeds, as relatively slow in ultrasound videos. We utilize expansion factors in both temporal and spatial domains (\ie, $\mathrm{\gamma_t}$ and $\mathrm{\gamma_s}$). Random masking with ratio $p_\mathrm{mask}$ is then applied to grids of the shape $\mathrm{\lceil \frac{L_p}{\gamma_t} \rceil\times \lceil \frac{H_p}{\gamma_s} \rceil\times \lceil \frac{W_p}{\gamma_s} \rceil}$, resulting in the chained masking. \cref{fig:STC} illustrates different masking strategies, including our Spatial-Temporal Chained (STC) masking.

Larger $\mathrm{\gamma_t}$ and $\mathrm{\gamma_s}$ can make masked regions more clustered. Combined with an appropriate masking ratio $p_\mathrm{mask}$, this helps balance the difficulty of the pretext task, improving the learning of spatial-temporal relationships.

\paragraph{Efficient Mamba-3D encoder without masked tokens}
In existing designs of masked autoencoders, all tokens are treated as fixed 1D sequences after embedding, making it relatively straightforward to remove masked tokens from the encoder for efficient training. However, in Mamba-3D blocks, where the 3D structure is preserved and the scanning direction alternates, it would be computationally intensive to reinsert all unmasked tokens into the 3D space and collect them again after permuting dimensions.

To address this, we employ a lightweight method that directly transforms all unmasked tokens in their flattened form under different permutations of the 3D structure, achieving approximately 40\% acceleration during the pre-training stage (from 68 hours to 42 hours). Specifically, we use additional 1D sequences to temporarily store the index transformations $\Psi_{[\cdot]\rightarrow[\cdot]}$ between different scans within the current mini-batch, thereby avoiding returning to the larger 3D space for each operation.

Let $\mathbf{x}^{(i)}_{[\cdot]}$ represent the flattened sequence of all unmasked tokens (including EGTs) after being arranged in a specific 3D arrangement. $i$ denotes as the input of the corresponding forward and reverse scanning in the $i$-th encoder block $(i=0,1,\dots,N-1)$. Thus, the initial $\mathbf{x}^{(0)}_\mathrm{LHW}$ is defined as:
\begin{equation}
\begin{aligned}
\mathbf{x}^{(0)}_\mathrm{LHW} &= \text{Flatten}\left[\text{Unmasked}\left(\mathbf{V}^{(0)}_{\text{in}}, 
\mathbf{M}\right)\right]
\\
&\in\mathbb{R}^{\mathrm{D\times S_{unmasked}}}
\end{aligned}\label{eq:x_0_LHW}
\end{equation}
where $\mathbf{M}$ is the binary mask for the entire $\mathbf{V}_{\text{in}}$ cube, and $\mathrm{S_{unmasked}}$ is the length of the flattened sequence. Then we have efficient transformations between different $\mathbf{x}^{(i)}_{[\cdot]}$:
\begin{equation}
\begin{aligned}
\mathbf{x}^{(i)}_\mathrm{HWL} &= \Psi_\mathrm{LHW\rightarrow HWL}\left( \mathbf{x}^{(i)}_\mathrm{LHW} \right)
\\
\mathbf{x}^{(i)}_\mathrm{LWH} &= \Psi_\mathrm{HWL\rightarrow LWH}\left( \mathbf{x}^{(i)}_\mathrm{HWL} \right)
\\
\mathbf{x}^{(i+1)}_\mathrm{LHW} &= \Psi_\mathrm{LWH\rightarrow LHW}\left( \mathbf{x}^{(i)}_\mathrm{LWH} \right)
\end{aligned}\label{eq:x_transformations}
\end{equation}
The implementation for accelerating the pre-training process of the Mamba-3D encoder without masked tokens is provided in Algorithm s1 in the supplementary material.

\paragraph{Loss function}
A linear projection $g: \mathrm{D \rightarrow D^{\prime}}$ (where $\mathrm{D^{\prime}=C\times p_t\times p_s\times p_s}$) is directly applied to the decoder's output ${\mathbf{V}^{\prime}}^{(M-1)}_{\mathrm{out}}$  , producing the final prediction $\mathbf{V}_\text{pred}=g\left({\mathbf{V}^{\prime}}^{(M-1)}_{\mathrm{out}}\right)$. The loss is calculated as the mean squared error (MSE) between predicted and original masked patches:
\begin{equation}
    \mathrm{L_{MVM}}=\frac{1}{\mathrm{N_{masked}\cdot D^{\prime}}}\sum_{i=1}^{\mathrm{N_{masked}}}\sum_{j=1}^{\mathrm{D^{\prime}}}\bigg(\Big[{\mathbf{V}_\text{pred}}\Big]_{ij} - \Big[{\mathbf{V}_\text{patch}}\Big]_{ij}\bigg)^2
  \label{eq:L_MVM}
\end{equation}

\begin{figure*}[t]
  \centering
  \includegraphics[width=\textwidth, height=0.24\textheight]{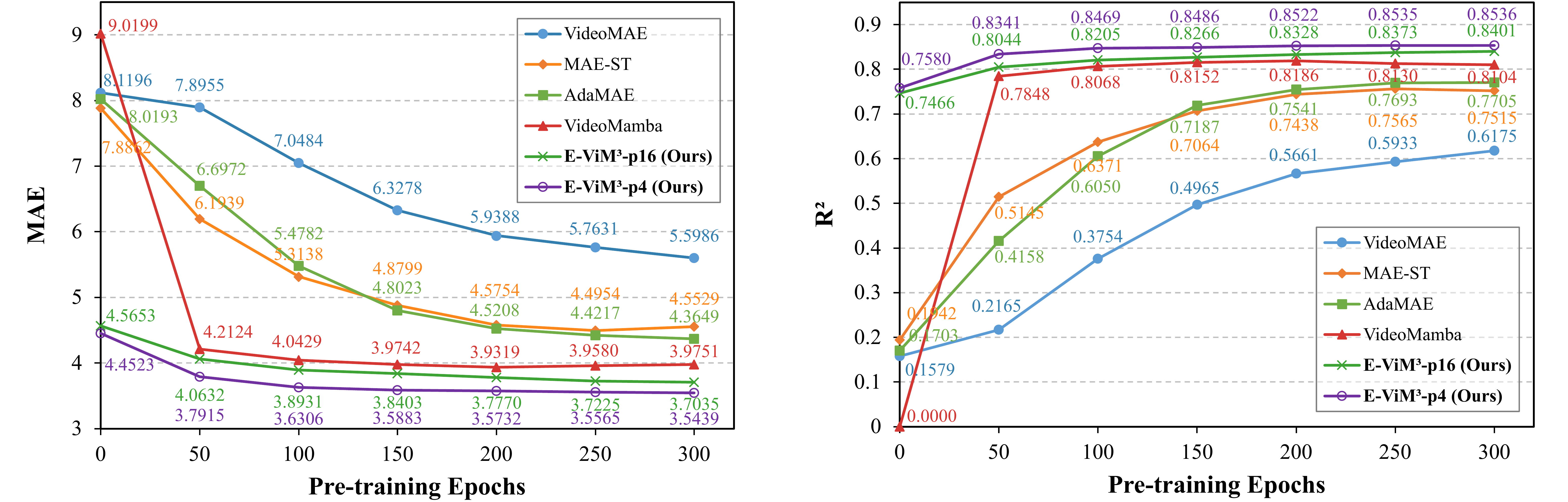}
  \vspace{-6mm}
  \caption{\textbf{Method comparisons and ablation studies with varying pre-training epochs on the EchoNet-Dynamic dataset.} The two key metrics both represent predictive accuracy. The horizontal axis at “0” indicates training from scratch on the downstream task.} 
  \label{fig:ablation_pretraining}
\end{figure*}

\begin{table*}[t]
\centering
\caption{\textbf{Results of the EF prediction task on the EchoNet-Dynamic dataset.} “SSP” denotes whether a self-supervised pre-training is used. All backbones for methods without “SSP” use checkpoints for video classification on Kinetics-400~\cite{Carreira1}. “F / P / C” refers to input frames, sampling period, and number of clips for inference. “Time” represents the inference wall time per sample on one RTX 4090. “\dag” denotes selecting video clips according to heart-beating cycles. “\ddag” denotes a reproduction of hyper-parameters not used in the original method. “$\ast$” indicates the default setting in other tables. The best and second-best results are \textbf{bolded} and \underline{underlined} respectively.}
\vspace{-8pt}
\belowrulesep=0pt
\aboverulesep=0pt
\footnotesize
\begin{tabular}{l|ccc|ccc|ccc}
\toprule
Method & SSP & Backbone & F / P / C & $\text{MAE}$$\downarrow$ & $\text{RMSE}$$\downarrow$ & $\text{R}^2$$\uparrow$ & Param (M) & Time (s)\\
\midrule
\multicolumn{9}{l}{\textit{Previous Methods}} \\
\hline
\quad EchoNet~\cite{Ouyang2} & \scalebox{0.8}{\XSolidBrush} & R(2+1)D-18 & 32 / 2 / all & 4.22 & 5.56 & 0.79 & 32 & - \\
\quad EchoNet~\cite{Ouyang2} & \scalebox{0.8}{\XSolidBrush} & R(2+1)D-18 & 32 / 2 / 5\textsuperscript{\dag} & 4.05 & 5.32 & 0.81 & 32 &  0.21 \\
\quad EchoCoTr~\cite{Muhtaseb1} & \scalebox{0.8}{\XSolidBrush} & UniFormer-S & 36 / 4 / 1 & 3.95 & 5.17 & 0.82 & 21 & 0.04 \\
\quad CoReEcho~\cite{Maani1} & \scalebox{0.8}{\XSolidBrush} & UniFormer-S & 36 / 4 / 3 & 3.90 & 5.13 & 0.82 & 21 & 0.11 \\
\quad EchoNarrator~\cite{Thomas2} & \scalebox{0.8}{\XSolidBrush} & ResNet-3D-18 & 16 / - / 1\textsuperscript{\dag} & 4.00 & - & - & 33 & 0.03 \\
\quad EchoMan~\cite{Lai1} & \scalebox{0.8}{\XSolidBrush} & R(2+1)D-18 & 32 / 2 / 1 & 3.93 & - & - & 32 & 0.05 \\
\quad CardiacNet~\cite{Yang2} & \scalebox{0.8}{\XSolidBrush} & ResNet-VQGAN & 16 / 3 / 1 & 3.83 & - & - & 28 & 4.55 \\
\hline
\quad VideoSwin~\cite{Liu2} & \scalebox{0.8}{\XSolidBrush} & Swin-S & 64 / 2 / 1 & 4.08 & 5.46 & 0.80 & 50 & 0.07 \\
\quad TimeSformer~\cite{Bertasius1} & \scalebox{0.8}{\XSolidBrush} & TimeSformer-384$\times$12 & 64 / 2 / 1 & 4.13 & 5.60 & 0.79 & 45 & 0.09 \\
\quad EchoNet\textsuperscript{\ddag} & \scalebox{0.8}{\XSolidBrush} & R(2+1)D-18 & 64 / 2 / 1 & 4.03 & 5.40 & 0.80 & 32 & 0.10 \\
\quad EchoCoTr\textsuperscript{\ddag} & \scalebox{0.8}{\XSolidBrush} & UniFormer-S & 64 / 2 / 1 & 3.95 & 5.24 & 0.82 & 21 & 0.05 \\
\quad CoReEcho\textsuperscript{\ddag} & \scalebox{0.8}{\XSolidBrush} & UniFormer-S & 64 / 2 / 1 & 3.92 & 5.15 & 0.82 & 21 & 0.05 \\

\hline
\multicolumn{9}{l}{\textit{Masked Video Modeling-Based Methods}} \\
\hline
\quad VideoMAE~\cite{Tong1} & \scalebox{0.8}{\Checkmark} & ViT-S & 64 / 2 / 1 & 5.60 & 7.56 & 0.62 & 22 & 0.06 \\
\quad MAE-ST~\cite{Feichtenhofer1} & \scalebox{0.8}{\Checkmark} & ViT-S & 64 / 2 / 1 & 4.55 & 6.10 & 0.75 & 22 & 0.06 \\
\quad AdaMAE~\cite{Bandara1} & \scalebox{0.8}{\Checkmark} & ViT-S & 64 / 2 / 1 & 4.36 & 5.89 & 0.77 & 22 & 0.06 \\
\quad VideoMamba\textsuperscript{$\ast$}~\cite{Li1} & \scalebox{0.8}{\Checkmark} & Bi-Mamba-384$\times$18 & 64 / 2 / 1 & 3.98 & 5.32 & 0.81 & 20 & 0.05 \\
\quad VideoMamba~\cite{Li1} & \scalebox{0.8}{\Checkmark} & Bi-Mamba-384$\times$36 & 64 / 2 / 1 & 3.95 & 5.25 & 0.82 & 38 & 0.09 \\
\color{gray!50}\quad Mamba-ND~\cite{Li2} w/o pre-training& \color{gray!50}\scalebox{0.8}{\XSolidBrush} & \color{gray!50}Mamba-3D-384$\times$6 & \color{gray!50}64 / 2 / 1 & \color{gray!50}5.37 & \color{gray!50}7.21 & \color{gray!50}0.65 & \color{gray!50}41 & \color{gray!50}0.08 \\
\quad Mamba-ND~\cite{Li2} w/ our SSP apporach & \scalebox{0.8}{\Checkmark} & Mamba-3D-384$\times$6 & 64 / 2 / 1 & 3.83 & 5.10 & 0.82 & 41 & 0.08 \\
\quad \textbf{E-ViM³-p16 (Ours)} & \scalebox{0.8}{\Checkmark} & Mamba-3D-384$\times$6 & 64 / 2 / 1 & \underline{3.70} & \underline{4.89} & \underline{0.84} & 41 & 0.09 \\
\quad \textbf{E-ViM³-p4 (Ours)} & \scalebox{0.8}{\Checkmark} & Mamba-3D-192$\times$6 & 64 / 2 / 1 & \textbf{3.54} & \textbf{4.68} & \textbf{0.85} & 12 & 0.13 \\
\bottomrule
\end{tabular}
\label{tab:echonet_main}
\end{table*}
\vspace{-8pt}


\subsection{\label{methods_sub3}Configurations}
\paragraph{Main hyper-parameters}
We adopt all default settings for Mamba layers as in the original Mamba~\cite{Dao1}. We set the default patch size to $\rm p_t=1$, $\rm p_s=16$ as E-ViM³-p16, with the model dimension $\mathrm{D}=384$ and encoder blocks $\mathrm{N}=6$ (totaling 36 Mamba layers).
For EGTs, we set $\mathrm{L_g=H_g=W_g}=3$.
We also build a more fine-grained network with $\rm p_t=2$, $\rm p_s=4$, and $\mathrm{D}=192$ as E-ViM³-p4, which behaves as a more powerful model in some scenarios.

In the pre-training phase, we use $\mathrm{M}=2$ blocks in the decoder to enhance the encoder's feature extraction. The default masking ratio is $p_\mathrm{mask}=0.8$, while the masking chain parameters $\mathrm{\gamma_t}$ and $\mathrm{\gamma_s}$ vary depending on the dataset and patch size. Details regarding other hyper-parameters can be found in the supplementary material.

\paragraph{Mitigating overfitting}
To address overfitting in downstream tasks, particularly with Mamba as observed by~\cite{Liu3, Li1}, we use the exponential moving average (EMA) model~\cite{He2, Izmailov1, Huang3} for validation and apply early stopping during fine-tuning. All other self-supervised pre-training methods follow the same approach for fair comparison.

\paragraph{Why not Mamba2?}
We opt for Mamba~\cite{Dao1} instead of Mamba2~\cite{Gu1}. Although Mamba2 improves usability for longer sequences and higher dimensions, the original Mamba is more efficient for the scale of our network and hardware. Additionally, there is no essential difference in their ability to model sequential data.

\begin{table*}[t]
\centering
\caption{\textbf{Results of the EF prediction task on the CAMUS dataset.} “SSP” denotes whether a self-supervised pre-training is used. “Echo-P” denotes whether trained from the best model on EchoNet-Dynamic. “Echo-SSP” denotes whether pre-trained from the best self-supervised pre-trained model on EchoNet-Dynamic. All backbones without “SSP” and “Echo-P” use checkpoints for video classification on Kinetics-400~\cite{Carreira1}. All methods use 64 frames as one clip for inference. “\dag” denotes a large gap from the original results due to the difference in data partitioning. The best and second-best results are \textbf{bolded} and \underline{underlined} in both upper and lower sections, respectively.}
\vspace{-8pt}
\belowrulesep=0pt
\aboverulesep=0pt
\footnotesize
\begin{tabular}{l|cccc|ccc}
\toprule
Method & SSP & Echo-P & Echo-SSP & Backbone & $\text{MAE}$$\downarrow$ & $\text{RMSE}$$\downarrow$ & $\text{R}^2$$\uparrow$ \\
\midrule
\multicolumn{8}{l}{\textit{Directly Trained on CAMUS}} \\
\hline
\quad EchoNet~\cite{Ouyang2} & \scalebox{0.8}{\XSolidBrush} & \scalebox{0.8}{\XSolidBrush} & - & R(2+1)D-18 & 7.56 & \underline{9.39} & \underline{0.41} \\
\quad EchoCoTr~\cite{Muhtaseb1} & \scalebox{0.8}{\XSolidBrush} & \scalebox{0.8}{\XSolidBrush} & - & UniFormer-S & 8.89 & 10.96 & 0.20 \\
\quad CoReEcho\textsuperscript{\dag}~\cite{Maani1} & \scalebox{0.8}{\XSolidBrush} & \scalebox{0.8}{\XSolidBrush} & - & UniFormer-S & 8.53 & 10.89 & 0.21 \\
\hline
\quad VideoMAE~\cite{Tong1} & \scalebox{0.8}{\Checkmark} & - & \scalebox{0.8}{\XSolidBrush} & ViT-S  & 9.38 & 12.11 & 0.02 \\
\quad MAE-ST~\cite{Feichtenhofer1} & \scalebox{0.8}{\Checkmark} & - & \scalebox{0.8}{\XSolidBrush} & ViT-S & 8.92 & 11.38 & 0.13 \\
\quad AdaMAE~\cite{Bandara1} & \scalebox{0.8}{\Checkmark} & - & \scalebox{0.8}{\XSolidBrush} & ViT-S & 8.83 & 11.43 & 0.13 \\
\quad VideoMamba~\cite{Li1} & \scalebox{0.8}{\Checkmark} & - & \scalebox{0.8}{\XSolidBrush} & Bi-Mamba-384$\times$18 & 9.41 & 12.23 & 0.00 \\
\quad Mamba-ND~\cite{Li2} w/ our SSP & \scalebox{0.8}{\Checkmark} & - & \scalebox{0.8}{\XSolidBrush} & Mamba-3D-384$\times$6 & \underline{7.36} & 9.49 & 0.40 \\
\quad \textbf{E-ViM³-p16 (Ours)} & \scalebox{0.8}{\Checkmark} & - & \scalebox{0.8}{\XSolidBrush} & Mamba-3D-384$\times$6 & \textbf{6.87} & \textbf{9.22} & \textbf{0.43} \\
\hline

\multicolumn{8}{l}{\textit{Pre-trained on EchoNet-Dynamic}} \\
\hline
\quad EchoNet~\cite{Ouyang2} & \scalebox{0.8}{\XSolidBrush} & \scalebox{0.8}{\Checkmark} & - & R(2+1)D-18 & \underline{6.49} & \underline{8.18} & \underline{0.55} \\
\quad EchoCoTr~\cite{Muhtaseb1} & \scalebox{0.8}{\XSolidBrush} & \scalebox{0.8}{\Checkmark} & - & UniFormer-S & 7.03 & 9.05 & 0.45 \\
\quad CoReEcho\textsuperscript{\dag}~\cite{Maani1} & \scalebox{0.8}{\XSolidBrush} & \scalebox{0.8}{\Checkmark} & - & UniFormer-S & 6.70 & 8.56 & 0.51 \\

\hline
\quad VideoMAE~\cite{Tong1} & \scalebox{0.8}{\Checkmark} & - & \scalebox{0.8}{\Checkmark} & ViT-S & 8.72 & 10.79 & 0.22 \\
\quad MAE-ST~\cite{Feichtenhofer1} & \scalebox{0.8}{\Checkmark} & - & \scalebox{0.8}{\Checkmark} & ViT-S & 8.43 & 10.55 & 0.26 \\
\quad AdaMAE~\cite{Bandara1} & \scalebox{0.8}{\Checkmark} & - & \scalebox{0.8}{\Checkmark} & ViT-S & 8.16 & 10.04 & 0.33 \\
\quad VideoMamba~\cite{Li1} & \scalebox{0.8}{\Checkmark} & - & \scalebox{0.8}{\Checkmark} & Bi-Mamba-384$\times$18 & 7.55 & 10.47 & 0.27 \\
\quad Mamba-ND~\cite{Li2} w/ our SSP & \scalebox{0.8}{\Checkmark} & - & \scalebox{0.8}{\Checkmark} & Mamba-3D-384$\times$6 & 6.85 & 8.42 & 0.53 \\
\quad \textbf{E-ViM³-p16 (Ours)} & \scalebox{0.8}{\Checkmark} & - & \scalebox{0.8}{\Checkmark} & Mamba-3D-384$\times$6 & \textbf{5.88} & \textbf{7.42} & \textbf{0.63} \\
\bottomrule
\end{tabular}
\label{tab:camus_main}
\end{table*}
\vspace{-8pt}
\begin{table*}[t]
\centering
\caption{\textbf{Results of the breast cancer classification task on the MICCAI-BUV and WHBUS datasets.} “Sen.” for Sensitivity, “Spe.” for Specificity, “Acc.” for Accuracy. The best results are \textbf{bolded}.}
\vspace{-8pt}
\belowrulesep=0pt
\aboverulesep=0pt
\footnotesize
\begin{tabular}{l|ccccc|ccccc}
\toprule
\multirow{2}{*}{Method} & \multicolumn{5}{c|}{MICCAI-BUV} & \multicolumn{5}{c}{WHBUS}\\
\cline{2-11}
& $\text{AUC}$$\uparrow$ & $\text{F1}$$\uparrow$ & $\text{Sen.}$$\uparrow$ & $\text{Spe.}$$\uparrow$ & $\text{Acc.}$$\uparrow$ & $\text{AUC}$$\uparrow$ & $\text{F1}$$\uparrow$ & $\text{Sen.}$$\uparrow$ & $\text{Spe.}$$\uparrow$ & $\text{Acc.}$$\uparrow$\\
\midrule
\multicolumn{6}{l}{\textit{Previous Methods}} \\
\hline
\quad SlowFast~\cite{Feichtenhofer2} & 82.84 & 80.00 & 69.23 & 92.31 & 76.92 & 85.00 & 82.35 & 82.35 & 85.00 & 83.78  \\
\quad R(2+1)D~\cite{Tran1} & 84.30 & 83.51 & 79.33 & 78.95 & 79.20 & 79.41 & 75.67 & 82.35 & 70.00 & 75.68 \\
\quad TSM~\cite{Lin3} & 87.28 & 88.00 & 84.62 & 84.62 & 84.62 & 84.41 & 81.08 & 88.24 & 75.00 & 81.08 \\
\quad TimeSformer~\cite{Bertasius1} & 83.73 & 83.33 & 76.92 & 84.62 & 79.49 & 88.53 & 84.21 & \textbf{94.12} & 75.00 & 83.78 \\
\quad MViTv2~\cite{Li4} & 85.80 & 87.50 & 80.77 & 92.31 & 84.62 & 88.53 & 88.24 & 88.24 & 90.00 & 89.19 \\
\quad SAG-Net~\cite{Zhang1} & 92.60 & 89.80 & 84.62 & 92.31 & 87.18 & 89.71 & \textbf{91.43} & \textbf{94.12} & 90.00 & \textbf{91.89} \\
\hline
\multicolumn{6}{l}{\textit{Masked Video Modeling-Based Methods}} \\
\hline
\quad VideoMAE~\cite{Tong1} & 83.43 & 83.33 & 76.92 & 84.62 & 79.49 & 79.12 & 77.41 & 70.59 & 90.00 & 81.08 \\
\quad MAE-ST~\cite{Feichtenhofer1} & 85.21 & 85.11 & 76.92 & 92.31 & 82.05 & 84.26 & 81.25 & 76.47 & 90.00 & 83.78 \\
\quad AdaMAE~\cite{Bandara1} & 90.53 & 90.20 & \textbf{88.46} & 84.62 & 87.18 & 90.83 & 85.71 & 88.24 & 85.00 & 86.49 \\
\quad VideoMamba~\cite{Li1} & 83.43 & 85.71 & 80.77 & 84.62 & 82.05 & 85.74 & 80.00 & 82.35 & 80.00 & 81.08 \\
\quad Mamba-ND~\cite{Li2} w/ our SSP & 92.88 & 87.50 & 77.78 & \textbf{100.00} & 85.00 & 89.56 & 85.71 & 88.24 & 85.00 & 86.49 \\
\quad \textbf{E-ViM³-p16 (Ours)} & \textbf{93.79} & \textbf{93.88} & \textbf{88.46} & \textbf{100.00} & \textbf{92.31} & \textbf{93.82} & 87.50 & 82.35 & \textbf{95.00} & 89.19 \\
\bottomrule
\end{tabular}
\label{tab:buv_main}
\end{table*}
\vspace{-8pt}

\section{Experiments}
\label{sec:experiments}

We compare our method with previous works and other masked video modeling approaches in~\cref{experiments_sub1}. Ablation studies on key designs and parameters are in~\cref{experiments_sub2}. Additional information including datasets and more experiments are available in the supplementary material.

\subsection{\label{experiments_sub1}Main Results}

\paragraph{EF prediction on the EchoNet-Dynamic dataset}
In~\cref{tab:echonet_main}, we compare our method to the baseline set by~\cite{Ouyang2} and other recent advanced works~\cite{Muhtaseb1, Maani1, Yang2} on the EchoNet-Dynamic dataset. Results show that our method greatly outperforms previous state-of-the-art methods and can better process and utilize longer video clips. 

Advanced MVM approaches were also evaluated, including VideoMAE~\cite{Tong1}, MAE-ST~\cite{Feichtenhofer1}, AdaMAE~\cite{Bandara1}, and the latest VideoMamba~\cite{Li1}.
We observe that ViT-based approaches struggle with this task, primarily because Transformers lack basic inductive biases and require huge datasets to train effectively. VideoMAE~\cite{Tong1} almost fails, as its tube masking strategy assumes drastic frame-to-frame changes, which is inappropriate for this scenario. In contrast, VideoMamba~\cite{Li1} demonstrates commendable performance. However, our E-ViM³ surpasses these methods distinctly by preserving the video data structure and enhancing feature aggregation through Enclosure Global Tokens.

\paragraph{EF prediction on the CAMUS dataset}
In~\cref{tab:camus_main}, we present the EF prediction results on the CAMUS dataset. Given the relatively small size of CAMUS, we provide results from fine-tuned (transferred) models pre-trained on the EchoNet-Dynamic dataset, alongside those trained directly on CAMUS, where the baseline methods still use the corresponding public pre-trained backbone. Our approach remains competitive with no additional data and has great advantages in the transfer learning scenario.

\paragraph{Breast cancer classification on the MICCAI-BUV and WHBUS datasets}
We further evaluate our method on two smaller datasets of another high-level task, with results shown in~\cref{tab:buv_main}. The previous state-of-the-art method~\cite{Zhang1} uses ConvNeXt-T~\cite{Liu4} pre-trained on ImageNet-22k~\cite{Russakovsky1} as the backbone and a clip-weighting strategy tailored for breast ultrasound videos. However, with a more general design and no ImageNet-based pre-training, our approach achieves superior performance on most key metrics, including an overall lead on the MICCAI-BUV dataset.

\paragraph{Effectiveness on a small portion of labeled data}
We also investigate whether our pre-trained E-ViM³ model can efficiently reduce the dependence on large amounts of labeled data. The results on the EchoNet-Dynamic dataset are shown in~\cref{tab:echonet_part}. With only 10\% training set for fine-tuning, our method achieves performance close to the baseline, indicating that E-ViM³ can greatly reduce reliance on labeled data after MVM pre-training. Results with 1\% training set are available in the supplementary material.

\subsection{\label{experiments_sub2}Ablation Studies}
All ablation studies are conducted on EchoNet-Dynamic, as it is a reliable benchmark for medical ultrasound video analysis. We use E-ViM³-p16 as the default model. More results are available in the supplementary material.

\begin{table}
\centering
\caption{\textbf{Results of the EF prediction task with 10\% training set labels on the EchoNet-Dynamic dataset.} “F / P” refers to input frames and sampling period. “\ddag” denotes a reproduction of hyper-parameters not used in the original method. The best and second-best results are \textbf{bolded} and \underline{underlined}, respectively.}
\vspace{-8pt}
\belowrulesep=0pt
\aboverulesep=0pt
\footnotesize
\begin{tabular}{l|c|ccc}
\toprule
Method & F / P & $\text{MAE}$$\downarrow$ & $\text{RMSE}$$\downarrow$ & $\text{R}^2$$\uparrow$ \\
\midrule
EchoNet~\cite{Ouyang2} & 32 / 2 & 5.93 & 8.26 & 0.54 \\
EchoCoTr~\cite{Muhtaseb1} & 36 / 4 & 5.16 & 6.96 & 0.68 \\
CoReEcho~\cite{Maani1} & 36 / 4 & 4.76 & 6.36 & 0.73 \\
\hline
EchoNet\ddag & 64 / 2 & 5.77 & 8.05 & 0.57 \\
EchoCoTr\ddag & 64 / 2 & 5.28 & 7.22 & 0.65 \\
CoReEcho\ddag & 64 / 2 & 4.86 & 6.47 & 0.72 \\
\hline
VideoMAE~\cite{Tong1} & 64 / 2 & 7.38 & 10.11 & 0.32 \\
MAE-ST~\cite{Feichtenhofer1} & 64 / 2 & 6.02 & 8.18 & 0.55 \\
AdaMAE~\cite{Bandara1} & 64 / 2 & 5.95 & 8.06 & 0.57 \\
VideoMamba~\cite{Li1} & 64 / 2 & 4.77 & 6.41 & 0.73 \\
Mamba-ND~\cite{Li2} w/ our SSP & 64 / 2 & 4.51 & 5.97 & 0.76 \\
\textbf{E-ViM³-p16 (Ours)} & 64 / 2 & \underline{4.17} & \underline{5.57} & \underline{0.79} \\
\textbf{E-ViM³-p4 (Ours)} & 64 / 2 & \textbf{4.00} & \textbf{5.29} & \textbf{0.81} \\
\bottomrule
\end{tabular}
\label{tab:echonet_part}
\end{table}
\vspace{-8pt}

\paragraph{How does pre-training influence downstream performance?}
We present the ablation results and method comparisons with varying pre-training epochs in~\cref{fig:ablation_pretraining}.
Compared with Transformer-based MVM methods, Mamba-based ones have prominently better performance with the same amount of training. VideoMamba\cite{Li1} struggles to converge without pre-training, and its performance declines after 200 epochs. In contrast, our method benefits from enhanced inductive biases and feature aggregation to perform best, achieving competitive results even from scratch. Training for 300 epochs is chosen as a cost-effective solution, which takes less than one day in our experimental setting, as detailed in the supplementary material.

\paragraph{How to integrate effective global tokens into the Mamba-3D backbone?}
In~\cref{tab:ablation_egts}, we present results for different global token designs. As in Mamba-ND~\cite{Li2}, the model without global tokens still performs reasonably, but with much slower convergence. We attribute this to global pooling, which leads to final features heavily influenced by irrelevant information. Likewise, for the proposed Enclosure Global Tokens, adding more internal grids does not always improve feature aggregation. A moderate setting yields the best results, outperforming both the case without global tokens and the approach of head-and-tail tokens.

\paragraph{How does the masking strategy affect the effectiveness of pre-training?}
As shown in~\cref{tab:ablation_stc}, when $\mathrm{\gamma_s}$ and $\mathrm{\gamma_t}$ are set to specific values, the masking strategy turns into widely used approaches (\eg, completely random, tube, and frame masking). In echocardiography, where targets rarely move substantially, tube masking performs the worst. In contrast, random masking performs much better, highlighting the importance of choosing effective masking strategies for different scenarios. Ablation results demonstrate that the proposed Spatial-Temporal Chained masking offers a reasonable pretext task, leading to the best performance.

\begin{table}
\centering
\caption{\textbf{Ablation study on the global token design on the EchoNet-Dynamic dataset.} The \colorbox{gray!20}{gray} row indicates the default setting for E-ViM³-p16. The best results are \textbf{bolded}.}
\belowrulesep=0pt
\aboverulesep=0pt
\footnotesize
\begin{tabular}{l|cc}
\toprule
Global Tokens & $\text{MAE}$$\downarrow$ & $\text{R}^2$$\uparrow$ \\
\midrule
None (average-pooling as~\cite{Li2}) & 3.8302 & 0.8247 \\
Head \& Tail & 3.8519 & 0.8239 \\
Enclosure ($\mathrm{L_g=H_g=W_g}=2$) & 3.7856 & 0.8332 \\
\rowcolor{gray!20} Enclosure ($\mathrm{L_g=H_g=W_g}=3$) & \textbf{3.7035} & \textbf{0.8401} \\
Enclosure ($\mathrm{L_g=5, H_g=W_g}=3$)  & 3.8413 & 0.8257 \\
\bottomrule
\end{tabular}
\label{tab:ablation_egts}
\end{table}

\begin{table}
\centering
\caption{\textbf{Ablation study on the masking strategy on the EchoNet-Dynamic dataset.} The \colorbox{gray!20}{gray} row indicates the default setting for E-ViM³-p16. The best results are \textbf{bolded}.}
\belowrulesep=0pt
\aboverulesep=0pt
\footnotesize
\begin{tabular}{l|cc}
\toprule
Masking Chain ($\mathrm{\gamma_s}$ / $\mathrm{\gamma_t}$) & $\text{MAE}$$\downarrow$ & $\text{R}^2$$\uparrow$ \\
\midrule
1 / 1 (random) & 3.8302 & 0.8247 \\
1 / 2 & 3.7856 & 0.8326 \\
\rowcolor{gray!20} 1 / 4 & \textbf{3.7035} & \textbf{0.8401} \\
1 / 16 & 4.0238 & 0.8043 \\
1 / 64 (tube) & 4.1620 & 0.7934 \\
2 / 1 & 3.7844 & 0.8335 \\
2 / 2 & 3.7524 & 0.8352 \\
2 / 4 & 3.9150 & 0.8198 \\
7 / 1 (frame) & 3.9193 & 0.8241 \\
\bottomrule
\end{tabular}
\label{tab:ablation_stc}
\end{table}

\section{Conclusion}
\label{sec:conclusion}

In this study, we proposed E-ViM³, a new Mamba-based model designed for analyzing medical ultrasound videos, which can be efficiently pre-trained as a masked autoencoder using cached index transformations and its inherent induction for 3D data. 
We introduced Enclosure Global Tokens (EGT) for enhanced feature aggregation within the 3D-preserving architecture, along with Spatial-Temporal Chained (STC) masking for flexible spatial-temporal learning across different video scenarios.
E-ViM³ outperforms previous state-of-the-art methods by a substantial margin on EchoNet-Dynamic and demonstrates its effectiveness on limited labels or smaller datasets.
Given its versatility, our approach can be extended to broader 3D and higher-dimensional visual data applications.


\clearpage
\appendix

\maketitlesupplementary


We provide more details and results of our work in this supplementary as follows:
\begin{itemize}
\item~\cref{sec:s_preliminaries}: A brief introduction to the concept and details of selective state space models as preliminaries.
\item~\cref{sec:s_datasets}: A description of the datasets used in this work.
\item~\cref{sec:s_implementation_details}: Implementation details, including hardware and software environments, detailed architectures, data pre-processing methods, and hyper-parameters.
\item~\cref{sec:s_experiments}: Additional experiments and ablation studies on some other key hyper-parameters.
\item~\cref{sec:s_visualization}: Visualization and Qualitative Analysis.
\end{itemize}


\section{Preliminaries on Selective State Space Models}
\label{sec:s_preliminaries}
Selective state space models (S6), better known as the efficient implementation in Mamba~\cite{Gu1}, are essentially derived from the vanilla state space models (SSM) in continuous linear time-invariant (LTI) systems.

SSMs utilize N-dimensional latent states $\mathbf{h}(t)\in\mathbb{R}^{N}$ to establish a complex mapping relationship from the one-dimensional continuous input signal to its output as $x(t)\in\mathbb{R}\rightarrow y(t)\in\mathbb{R}$:
\begin{equation}
\begin{aligned}
\mathbf{h}^{\prime}(t)&=\mathbf{A}\mathbf{h}(t)+\mathbf{B}x(t) \\
y(t)&=\mathbf{C}\mathbf{h}(t)+\mathbf{D}x(t)
\end{aligned}\label{eq:ssm}
\end{equation}
where $\mathbf{A}\in\mathbb{R}^{N\times N}$ is the “state matrix”, $\mathbf{B}\in\mathbb{R}^{N\times 1}$ is the “input matrix”, $\mathbf{C}\in\mathbb{R}^{1\times N}$ is the “output matrix”, $\mathbf{D}\in\mathbb{R}^{1\times 1}$ is the “feed-through matrix” as a direct shortcut from inputs to outputs, and the derivative $\mathbf{h}^{\prime}(t):=\frac{d}{dt}\mathbf{h}(t)$ represents the changing rate of latent states.

To apply to deep neural networks where data and parameters are mostly discretized, structured state space sequence models (S4)~\cite{Gu2} explicitly discretize the parameter matrices $\mathbf{A}$ and $\mathbf{B}$ in~\cref{eq:ssm} according to specific methods (\eg, Euler, bilinear or zero-order hold)~\cite{Smith1}. Zero-order hold (ZOH) is introduced here as utilized in Mamba~\cite{Gu1}:
\begin{equation}
\begin{aligned}
\overline{\mathbf{A}}&=\exp(\Delta \mathbf{A})\\
\overline{\mathbf{B}}&=(\Delta \mathbf{A})^{-1}\left[\exp(\Delta \mathbf{A})-\mathbf{I}\right]\cdot\Delta \mathbf{B}\approx\Delta \mathbf{B}
\end{aligned}\label{eq:zoh}
\end{equation}
where parameter $\Delta$ indicates the step size and $\mathbf{A}$ is a diagonal in Mamba, which makes $(\Delta \mathbf{A})^{-1}\left[\exp(\Delta \mathbf{A})-\mathbf{I}\right]\approx\mathbf{I}$ when all the entries in $\Delta \mathbf{A}$ are small enough, Therefore, S4 can be applied to one-dimensional sequential data as $x_t\in\mathbb{R}\rightarrow y_t\in\mathbb{R}$:
\begin{equation}
\begin{aligned}
\mathbf{h}_{t}&=\overline{\mathbf{A}}\mathbf{h}_{t-1}+\overline{\mathbf{B}}x_{t}\\
y_{t}&=\mathbf{C}\mathbf{h}_{t}+\mathbf{D}x_{t}
\end{aligned}\label{eq:s4}
\end{equation}
For high-dimensional input and output sequences (\ie, more than one channel), the above parameters and operations are independent in each channel.

Based on the above, the S6 structure used by Mamba~\cite{Gu1} further improves the capacity of the model by making $\overline{\mathbf{A}}$, $\overline{\mathbf{B}}$, $\mathbf{C}$ input-dependent across all channels of $\mathbf{x}_t\in\mathbb{R}^{D}$, which is known as the selection mechanism:
\begin{equation}
\begin{aligned}
\mathbf{B}_t&=S_B\left(\mathbf{x}_t\right)=\mathrm{Linear}_{D\rightarrow N}\left(\mathbf{x}_t\right) \\
\mathbf{C}_t&=S_C\left(\mathbf{x}_t\right)=\mathrm{Linear}_{D\rightarrow N}\left(\mathbf{x}_t\right) \\
\mathbf{\Delta}_t&=\tau_{\Delta}\left[S_\Delta\left(\mathbf{x}_t\right)\right]=\tau_{\Delta}\left[\mathrm{Linear}_{D\rightarrow D}\left(\mathbf{x}_t\right)\right] \\
\end{aligned}\label{eq:s6}
\end{equation}
where $\tau_{\Delta}=\mathrm{softplus}$. With the discretization in~\cref{eq:zoh}, $\overline{\mathbf{A}}$, $\overline{\mathbf{B}}$, $\mathbf{C}$ are no longer fixed parameters in the inference phase, making the system much more complex and more expressive than the original SSMs.



\section{Datasets}
\label{sec:s_datasets}

\paragraph{EchoNet-Dynamic}
EchoNet-Dynamic~\cite{Ouyang1, Ouyang2} is a public echocardiographic dataset consisting of 7,460 apical four-chamber (A4C) videos for training, 1,288 for validation, and 1,276 for testing, with a varying length from dozens to thousands of 112x112 grayscale frames. Following previous methods~\cite{Muhtaseb1, Maani1}, we only use ejection fraction (EF) labels for end-to-end learning with MSE loss and follow the augmentations made by the original baseline EchoNet~\cite{Ouyang2}. We use mean absolute error ($\text{MAE})$ and coefficient of determination ($\text{R}^2$) as performance metrics, supplemented by root mean squared error ($\text{RMSE}$).

\paragraph{CAMUS}
CAMUS~\cite{Leclerc1} is another public echocardiogram dataset with video data and annotations from 500 patients for ejection fraction (EF) prediction. We follow the partitioning given by the dataset provider, as the ratio of training-validation-testing is 8:1:1. Unlike EchoNet-Dynamic~\cite{Ouyang1} with videos of several cardiac cycles, each video in CAMUS consists of a single heartbeat in both apical two-chamber (A2C) and apical four-chamber (A4C) views. We use A4C videos for end-to-end EF prediction. Following~\cite{Maani1}, we no longer directly sample frames as in EchoNet-Dynamic but convert each video to the desired resolution and length by trilinear interpolation. The loss function and metrics are the same as for EchoNet-Dynamic.
      
\paragraph{MICCAI-BUV}
MICCAI-BUV~\cite{Lin2} is a public breast ultrasound video dataset with a complete scan of the tumor. Compared to EchoNet-Dynamic~\cite{Ouyang1}, MICCAI-BUV consists of 114 malignant samples and 71 benign ones. The authors of the dataset randomly and evenly selected 20\% to form a validation set. This dataset contains tumor detection bounding boxes and binary labels, and we only use the binary labels for classification to compare to the existing method~\cite{Zhang1}. We use weighted BCE loss and classification evaluation metrics including AUC (Area Under the ROC Curve), F1-score, sensitivity, specificity, and accuracy.

\paragraph{WHBUS}
WHBUS~\cite{Zhang1} is another public breast ultrasound video dataset besides MICCAI-BUV. WHBUS
contains 83 malignant samples and 101 benign ones, which have also been randomly and evenly divided into 8:2 as the training and validation sets. We also use weighted BCE loss and classification evaluation metrics as for the MICCAI-BUV dataset. 

For MICCAI-BUV and WHBUS which are quite small, we employ more data augmentation techniques than the previous state-of-the-art method~\cite{Zhang1} to compensate for not using any additional data (\eg, ImageNet-22k) to pre-train the backbone. Details are available in~\cref{sec:s_implementation_details}.


\section{Implementation Details}
\label{sec:s_implementation_details}
We conducted all experiments in a Linux environment using a maximum of two NVIDIA RTX 4090 GPUs during both the pre-training and fine-tuning phases, with all inferences performed on a single GPU. We use Python (version 3.10) and PyTorch~\cite{Paszke1} (version 2.3.1) along with other necessary common packages. Distributed-Data-Parallel is adopted in training phases. In cases where the required VRAM exceeded available capacity, we utilized gradient accumulation without any adverse effects, as operations sensitive to this process, such as BatchNorm, were not included in our network architecture.

In~\cref{alg:Eff}, we show the detailed implementation for speeding up the Mamba-3D encoder without masked tokens in the form of PyTorch-style pseudocode.

\begin{figure}[H]
  \centering
  \includegraphics[width=\columnwidth,keepaspectratio]{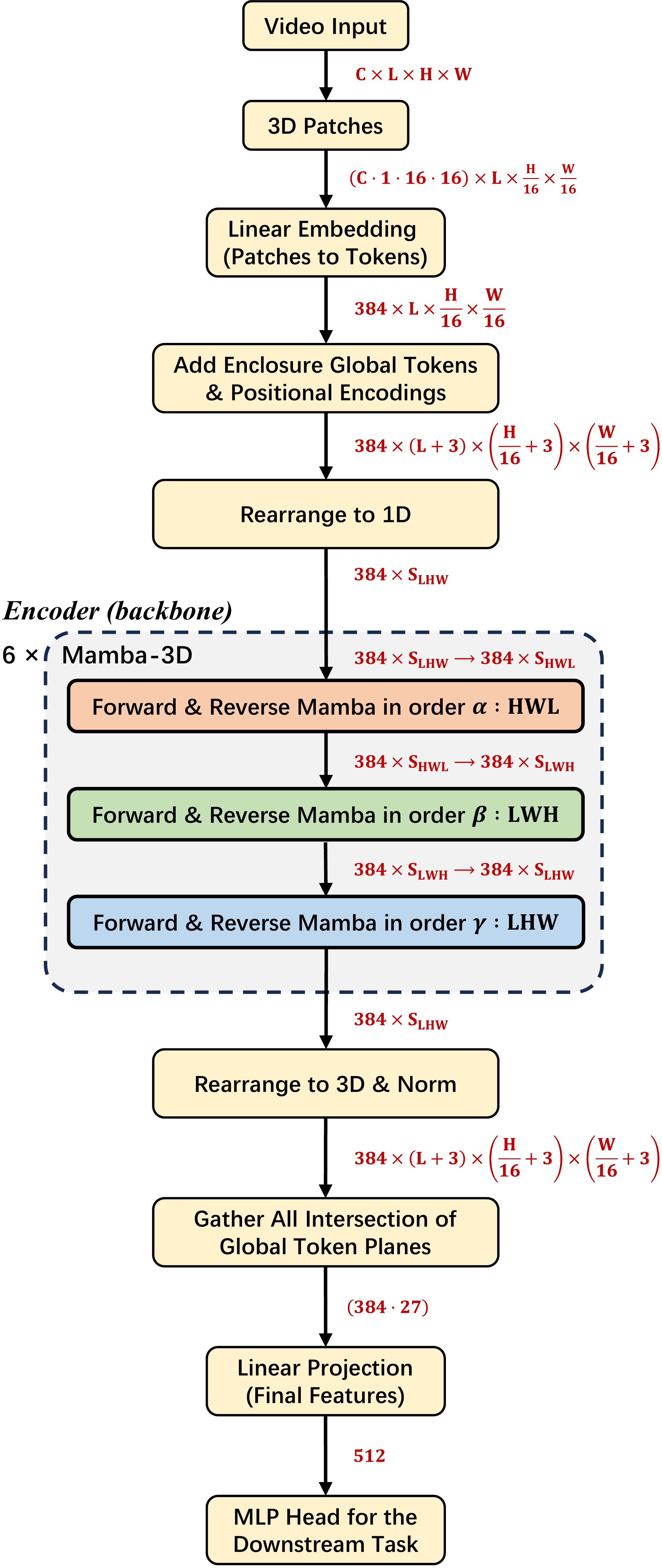}
  \caption{\textbf{Overall architecture of the default E-ViM³-p16 model for downstream tasks.} The details of the encoder (backbone) are further illustrated in~\cref{fig:overall_pretrain}.}
  \label{fig:overall_downstream}
\end{figure}

\begin{figure*}[t]
  \centering
  \includegraphics[width=\textwidth,height=\textheight,keepaspectratio]{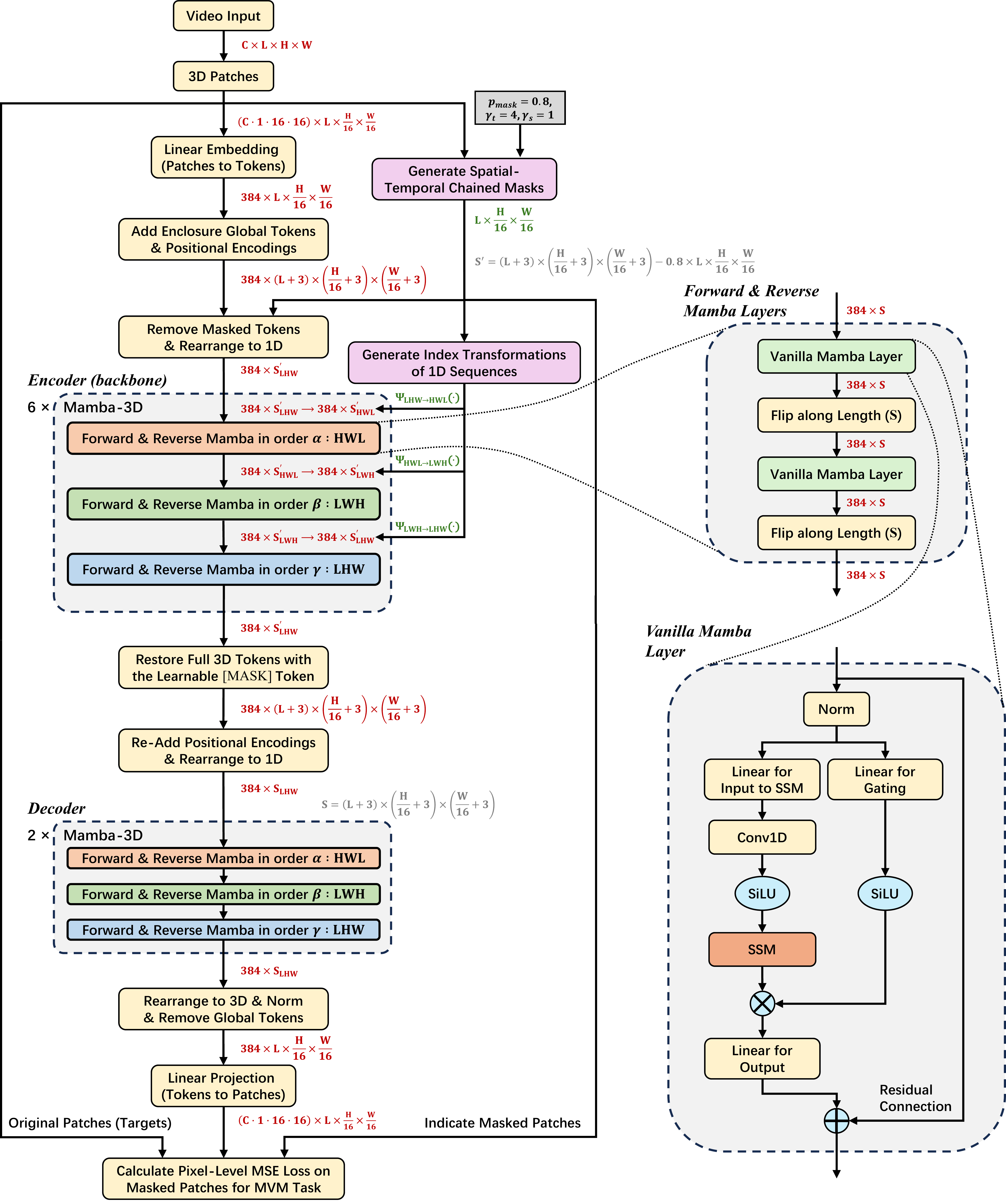}
  \caption{\textbf{Overall architecture of the default E-ViM³-p16 model as a masked autoencoder for self-supervised pre-training.} For details on the State Space Model (SSM) module in the vanilla Mamba layer, please refer to~\cite{Gu1}, or get a brief overview in~\cref{sec:s_preliminaries}.}
  \label{fig:overall_pretrain}
\end{figure*}

In~\cref{fig:overall_downstream} and~\cref{fig:overall_pretrain}, we provide detailed network architecture of the default E-ViM³-p16 model for the stages of downstream fine-tuning and self-supervised pre-training, respectively, as a supplement to Figure 1 in the main paper.

In~\cref{tab:default_preprocessing}, we detail the properties of each dataset, including pre-processing and data augmentation techniques employed.

We list the default configurations and hyper-parameters of E-ViM³-p16 for pre-training and fine-tuning In~\cref{tab:default_pretrain_configs} and~\cref{tab:default_finetune_configs}, respectively. For the additional E-ViM³-p4 model used on the Echonet-Dynamic dataset, all the changed hyper-parameters are clearly shown in~\cref{tab:ablation_patch_size_dim}.


\section{Additional Experimental Results}
\label{sec:s_experiments}

\paragraph{Segmentation task with non-global tokens}
We also evaluated E-ViM³ on the left ventricular segmentation task of EchoNet-Dynamic. As shown in~\cref{tab:s_seg}, our method has good segmentation performance at both end-diastolic (ED) and end-systolic (ES) phases, with E-ViM³-p4 even surpassing state-of-the-art methods that are specifically tailored for this task.

\paragraph{Effectiveness on 1\% labeled data}
In~\cref{tab:s_echonet_part}, we show the fine-tuned results on the EchoNet-Dynamic dataset with only 1\% training set labels. The methods without self-supervised pre-training all failed, along with the two ViT-based pre-training methods. However, E-ViM³ is still effective and significantly better than VideoMamba~\cite{Li1}. This further reflects the advantage of E-ViM³ in inductive bias, which makes it easier to be effectively trained.

\paragraph{Ablation on the masking ratio}
In~\cref{tab:ablation_ratio}, we show the impact of different masking ratios. As with existing leading masked video modeling methods~\cite{Tong1, Feichtenhofer1, Li1}, a slightly larger mask ratio is still the key to efficient learning. It may vary from dataset to dataset, but we take $p_\mathrm{mask}=0.8$ as a more general and reasonable choice for all experiments.

\paragraph{Ablation on the input length and sampling strategy}
In this study, we only change the number of frames and the sampling period. Considering the limited number of frames in the EchoNet-Dynamic dataset, we tested some reasonable combinations as shown in~\cref{tab:ablation_frames}. 

The optimum can be achieved with up to 64 frames and a sampling period of 2, while further increasing the number of frames brings no better results. This may stem from the redundant information caused by the periodic repetition of echocardiographic data, and being too long makes it harder to model. However, with more inner global tokens on the frame dimension (\ie, $\mathrm{L_g}$), the performance also improves, showing the possibility of further expansion.

\paragraph{Ablation on different patch-size and embedding dimensions}
In~\cref{tab:ablation_patch_size_dim}, we show the results of two different video embedding methods (\ie, different spatial and temporal patch-size combinations $\mathrm{p_s}$, $\mathrm{p_t}$) with different embedding dimensions (\ie, model dimensions $\mathrm{D}$).

When using $\mathrm{p_s=16}$ (\ie, E-ViM³-p16), the moderate model dimension ($\mathrm{D=384}$) achieves better performance, while increasing it further leads to more overfitting on the EchoNet-Dynamic dataset due to excessive complexity. For patches that are smaller in space and slightly larger in time as $\mathrm{p_s=4}$ and $\mathrm{p_t=2}$ (\ie, E-ViM³-p4), the model achieves better performance with even fewer parameters.

\paragraph{Ablation on the number of encoder blocks}
In~\cref{tab:ablation_encoder_block_number}, we show the results of E-ViM³-p16 with different numbers of encoder blocks (\ie, $\mathrm{N}$).

In particular, $\mathrm{N}=3$ is the E-ViM³-p16-S in the main paper Table 1, and $\mathrm{N}=6$ is the standard E-VIM³-p16. When further increasing $\mathrm{N}$ to 12, the model is still data-efficient. However, considering that doubling the depth does not bring enough improvement, we still use $\mathrm{N}=6$.


\section{Visualization and Qualitative Analysis}
\label{sec:s_visualization}

\begin{figure*}[t]
  \centering
  \includegraphics[width=\textwidth,height=\dimexpr\textheight-2.0cm\relax,keepaspectratio]{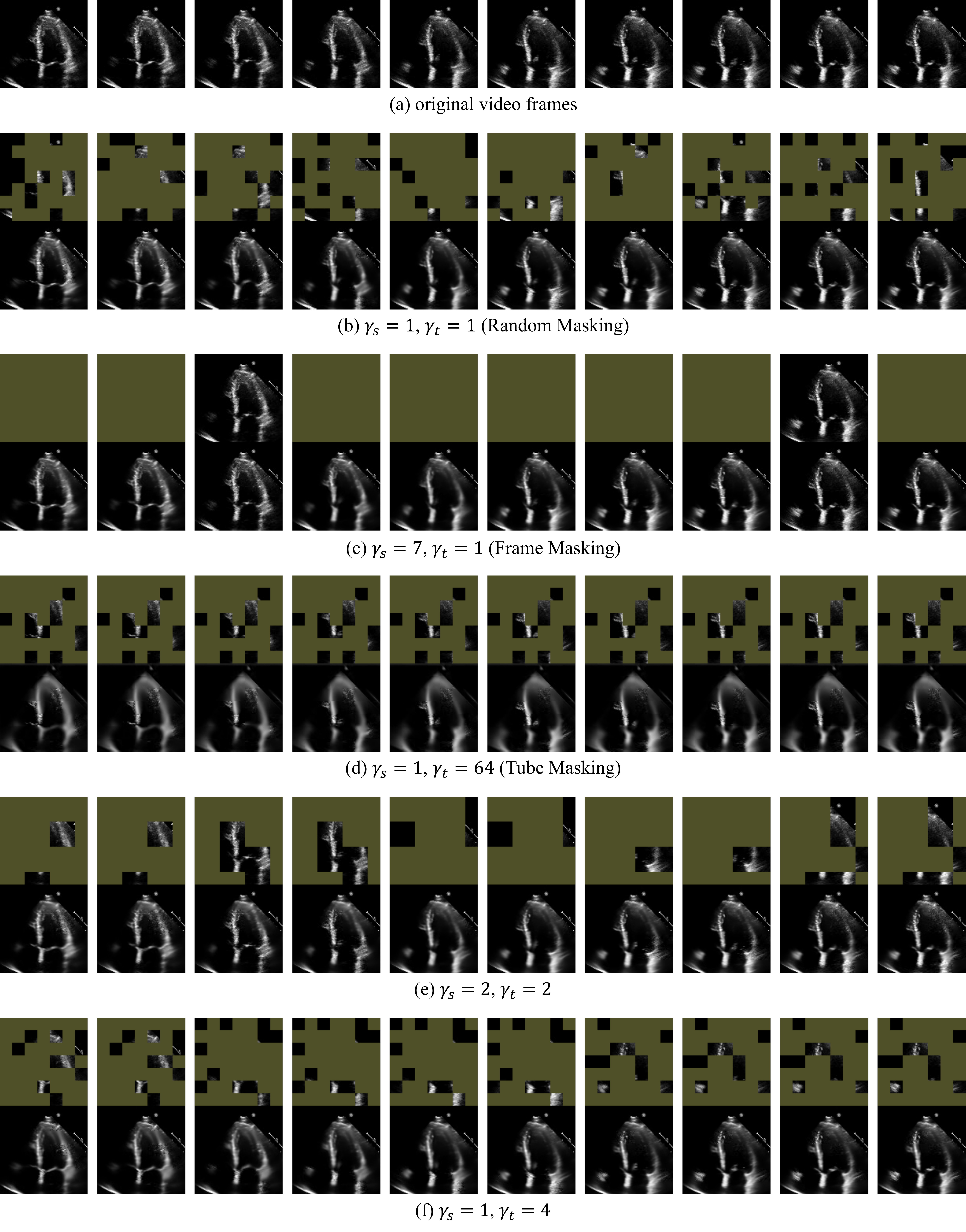}
  \caption{\textbf{Visualization of different mask strategy and their corresponding reconstruction results of our E-ViM³-p16.} Each row consists of 10 consecutive frames from left to right. In this task, the optimal balance point for masking is to avoid giving away too much information while making the model sufficient to recover the general dynamic process. This sample is “0X10A28877E97DF540” in the test set of EchoNet-Dynamic.}
  \label{fig:mask_examples}
\end{figure*}

\paragraph{Different masking strategies and reconstruction results}
In~\cref{fig:mask_examples}, we visualize different masking strategies discussed in this paper and present the reconstruction results given by our E-ViM³-p16 model after pre-training. Intuitively, the desired effect is that the masked video does not provide too many visual features brought by heartbeats, but the reconstructed video recovers these as much as possible.

\begin{figure*}[t]
  \centering
  \includegraphics[width=\textwidth,height=\dimexpr\textheight-2.0cm\relax,keepaspectratio]{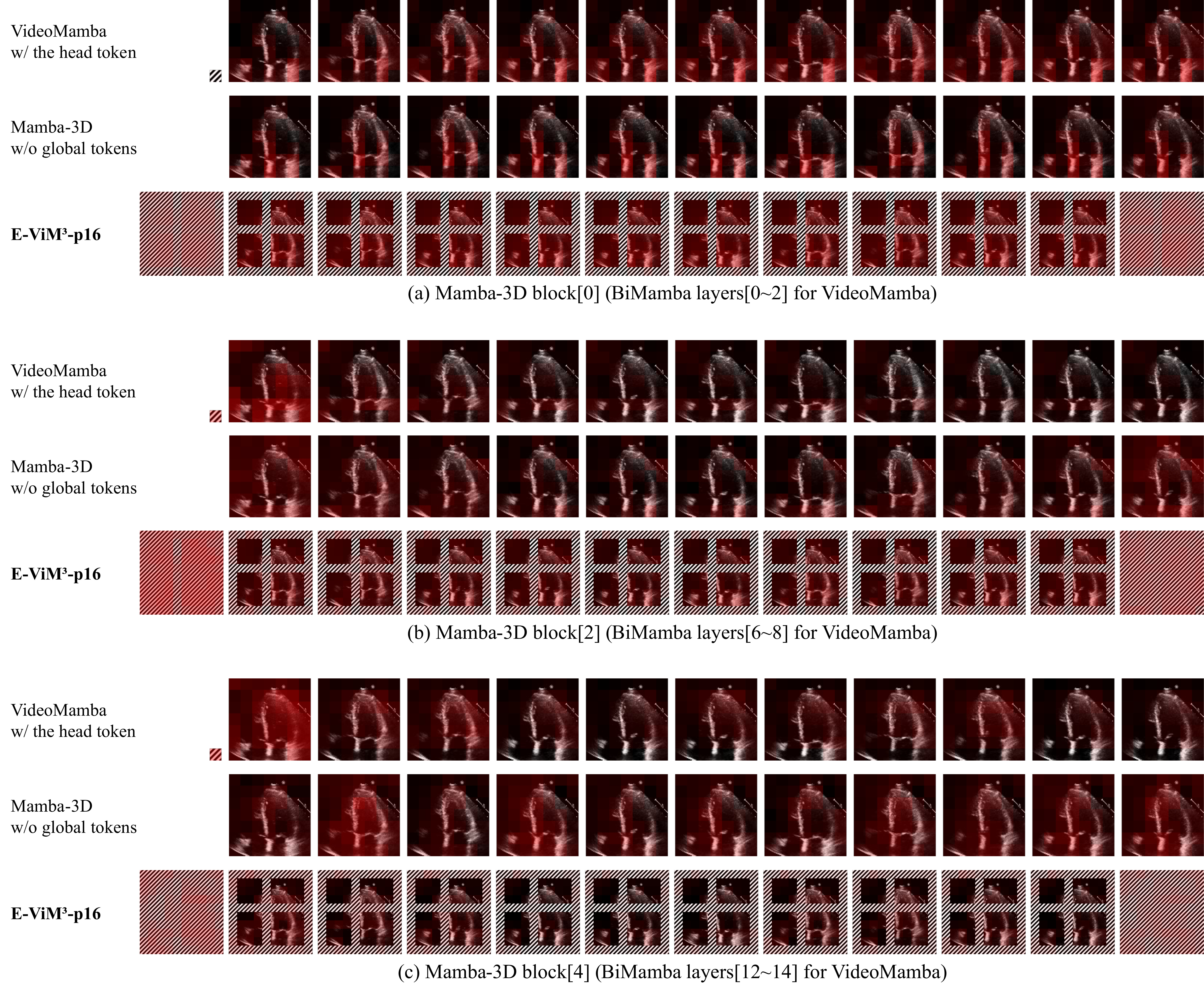}
  \caption{\textbf{Visualization of Delta ($\mathbf{\Delta}_t$) in different Mamba-based video models.} The compared models: VideoMamba, vanilla Mamba-3D without global tokens, and E-ViM³-p16, all consist of 36 SSM modules. We take every 6 of them as a group, which is exactly one Mamba-3D block or three BiMamba layers. All channels of grouped $\mathbf{\Delta}_t$ are concatenated, averaged, and finally normalized for visualization. The red mask in the figure indicates the numerical relationship of Delta, which from light to deep represents the normalized value of 0 to 1. As shown in the figure, with the increasing depth of the network, E-ViM³ demonstrates a superior ability to focus on relevant information and effectively aggregate features by Enclosure Global Tokens. This sample is “0X10A28877E97DF540” in the test set of EchoNet-Dynamic.}
  \label{fig:delta_visualization}
\end{figure*}

\paragraph{Delta Visualizations for different Mamba-based models}
In~\cref{fig:delta_visualization}, we visualize the input-dependent Delta ($\mathbf{\Delta}_t$) in VideoMamba~\cite{Li1}, Mamba-3D without global tokens~\cite{Li2} and E-ViM³-p16. As a crucial parameter in Mamba layers, it plays an important role in selective scanning. Its non-negative values are similar to attention weights and the visualization allows for a more intuitive understanding of the model's focus on different regions of input data.

\clearpage


\begin{table}[t]
\centering
\caption{\textbf{Results of the left ventricular segmentation task on EchoNet-Dynamic.} The best results are \textbf{bolded}.}
\belowrulesep=0pt
\aboverulesep=0pt
\small
\setlength{\tabcolsep}{5pt}
\begin{tabular}{l|ccc}
\hline
Method & Dice (ES) & Dice (ED) & Dice (overall) \\
\hline
EchoNet-Seg~\cite{Ouyang2}    & 0.9068 & 0.9278 & 0.9200 \\
nnUNet~\cite{nnunet}         & 0.9163 & 0.9362 & 0.9286 \\
SepXception~\cite{SepXception}    & 0.9173 & 0.9364 & 0.9290 \\
SimLVSeg-SI~\cite{SimLVSeg}    & 0.9226 & 0.9395 & 0.9331 \\
SimLVSeg-3D~\cite{SimLVSeg}    & 0.9229 & 0.9395 & 0.9332 \\
\textbf{E-ViM³-p16 (Ours)}  & 0.9209 & 0.9394 & 0.9324 \\
\textbf{E-ViM³-p4 (Ours)}   & \textbf{0.9247} & \textbf{0.9420} & \textbf{0.9354} \\
\hline
\end{tabular}
\label{tab:s_seg}
\end{table}

\begin{algorithm}[t]
\caption{Pseudocode of the efficient encoder in MVM pre-training for E-ViM³.}
\label{alg:Eff}
\definecolor{codeblue}{rgb}{0.25,0.5,0.5}
\lstset{
  backgroundcolor=\color{white},
  basicstyle=\fontsize{7.2pt}{7.2pt}\ttfamily\selectfont,
  columns=fullflexible,
  breaklines=true,
  captionpos=b,
  commentstyle=\fontsize{7.2pt}{7.2pt}\color{codeblue},
  stringstyle=\fontsize{7.2pt}{7.2pt}\color{codeblue},
  keywordstyle=\fontsize{7.2pt}{7.2pt},
  showstringspaces=false,
  showspaces=false,
  showtabs=false,
}
\begin{lstlisting}[language=python]
'''
"N" as batch-size
Inputs:
    v [N, L, H, W, D]: embedding with global tokens
    mask [N, L, H, W]: boolean mask, True for masked
Modules:
    encoder_blocks (forward and reverse of block i):
        f_hwl_i, r_hwl_i: Scans for HWL
        f_lwh_i, r_lwh_i: Scans for LWH
        f_lhw_i, r_lhw_i: Scans for LHW
Main intermediate results:
    lhw_2_hwl [uncertain, D]: LHW-1D -> HWL-1D
    hwl_2_lwh [uncertain, D]: HWL-1D -> LWH-1D
    lwh_2_lhw [uncertain, D]: LWH-1D -> LHW-1D
Outputs:
    v [N, L, H, W, D]
'''
unmasked_arange = torch.arange(~mask.sum())

lhw_order = -torch.ones_like(mask, dtype=torch.long)
lhw_order[~mask] = unmasked_arange
lhw_order = rearrange(lhw_order, 'NLHW->NHWL')
mask = rearrange(mask, 'NLHW->NHWL')
lhw_2_hwl = lhw_order[~mask]

hwl_order = lhw_order
hwl_order[~mask] = unmasked_arange
hwl_order = rearrange(hwl_order, 'NHWL->NLWH')
mask = rearrange(mask, 'NHWL->NLWH')
hwl_2_lwh = hwl_order[~mask]

lwh_order = hwl_order
lwh_order[~mask] = unmasked_arange
lwh_order = rearrange(lwh_order, 'NLWH->NLHW')
mask = rearrange(mask, 'NLWH->NLHW')
lwh_2_lhw = lwh_order[~mask]

x = v[~mask].reshape(N, -1, D)  # x_lhw
N, S, D = x.shape
for encoder_block_i in encoder_blocks:
    # 1D sequence rearrangement: LHW -> HWL
    x = x.flatten(0, 1)[lhw_2_hwl].reshape(N, S, D)
    x = r_hwl_i(f_hwl_i(x).flip(1)).flip(1)
    
    # 1D sequence rearrangement: HWL -> LWH
    x = x.flatten(0, 1)[hwl_2_lwh].reshape(N, S, D)
    x = r_lwh_i(f_lwh_i(x).flip(1)).flip(1)

    # 1D sequence rearrangement: LWH -> LHW
    x = x.flatten(0, 1)[lwh_2_lhw].reshape(N, S, D)
    x = r_lhw_i(f_lhw_i(x).flip(1)).flip(1)
v[~mask] = x.flatten(0, 1)
return v
\end{lstlisting}
\end{algorithm}

\clearpage

\begin{table*}[t]
\centering
\caption{\textbf{Detailed properties including pre-processing and augmentation procedures for all datasets.}}
\belowrulesep=0pt
\aboverulesep=0pt
\small
\begin{tabular}{l|cccc}
\toprule
\hline
\multirow{2}{*}{Property} & \multicolumn{4}{c}{Datasets} \\
\cline{2-5}
& EchoNet-Dynamic & CAMUS & MICCAI-BUV & WHBUS \\
\midrule
total videos & 10k & 500 & 185 & 184 \\
video length & 28 $\sim$ 1002 & 10 $\sim$ 42 & 30 $\sim$ 390 & 24 $\sim$ 457 \\
frame resolution & 112 × 112 & \textgreater\ 292 × 323 & \textgreater\ 442 × 442 & 384 × 450 \\
task type & numerical regression & numerical regression & binary classification & binary classification \\
\hline
pre-processing & \makecell[l]{1. normalization} & \makecell[l]{1. trilinear interpolation \\ \quad to 64 × 112 × 112 \\ 2. normalization} & \makecell[l]{1. spatially resizing \\ \quad to 224 × 224 \\ 2. normalization \\ \textit{3. resizing to 112 × 112} \\ \textit{\quad when not training}} & \makecell[l]{1. spatially resizing \\ \quad to 224 × 224 \\ 2. normalization \\ \textit{3. resizing to 112 × 112} \\ \textit{\quad when not training}} \\
\hline
augmentation & \makecell[l]{1. random sampling \\ 2. zero-padding \\ \quad to 136 × 136 \\ 3. random cropping \\ \quad to 112 × 112} & \makecell[l]{1. rotation ($\pm$10°) \\ 2. scaling (0.8 $\sim$ 1.1) \\ 3. translation (10\%)} & 
\makecell[l]{1. random sampling \\ 2. rotation ($\pm$30°) \\ 3. scaling (0.8 $\sim$ 1.2) \\ 4. translation (20\%) \\ 5. resized cropping \\ \quad to 112 × 112 \\ \quad with scale (0.8 $\sim$ 1.0) \\ \quad and ratio (3/4 $\sim$ 4/3) \\ 6. color jitter \\ \quad with brightness (0.2) \\ \quad and contrast (0.2) \\ 7. horizontal flip (50\%) \\ 8. \textit{mix-up with $\alpha=1.0$} \\ \textit{\quad when pre-training}} & \makecell[l]{1. random sampling \\ 2. rotation ($\pm$30°) \\ 3. scaling (0.8 $\sim$ 1.2) \\ 4. translation (20\%) \\ 5. resized cropping \\ \quad to 112 × 112 \\ \quad with scale (0.8 $\sim$ 1.0) \\ \quad and ratio (3/4 $\sim$ 4/3) \\ 6. color jitter \\ \quad with brightness (0.2) \\ \quad and contrast (0.2) \\ 7. horizontal flip (50\%) \\ 8. \textit{mix-up with $\alpha=1.0$} \\ \textit{\quad when pre-training}} \\
\bottomrule
\end{tabular}
\label{tab:default_preprocessing}
\end{table*}


\begin{table*}[t]
\centering
\caption{\textbf{Default configurations and hyper-parameters of E-ViM³-p16 for self-supervised pre-training.}}
\belowrulesep=0pt
\aboverulesep=0pt
\small
\begin{tabular}{l|cccc}
\toprule
\hline
\multirow{2}{*}{Configuration \& Hyper-parameter} & \multicolumn{4}{c}{Datasets} \\
\cline{2-5}
& EchoNet-Dynamic & CAMUS & MICCAI-BUV & WHBUS \\
\midrule
optimizer & \multicolumn{4}{c}{AdamW~\cite{Loshchilov1} with $\beta_1=0.9, \beta_2=0.999, \varepsilon=1\mathrm{e}{-15}$} \\
weight decay & \multicolumn{4}{c}{$1\mathrm{e}{-1}$} \\
learning rate schedule & \multicolumn{4}{c}{cosine decay with linear warm-up} \\
basic learning rate & \multicolumn{4}{c}{$1\mathrm{e}{-3}$} \\
minimal learning rate & \multicolumn{4}{c}{$1\mathrm{e}{-5}$} \\
warm-up steps & 5 epochs & 500 steps & 500 steps & 500 steps \\
total steps & 300 epochs & 10k steps & 20k steps & 20k steps \\
total batch size & \multicolumn{4}{c}{$32=4(\text{batch size per rank}) \times 2(\text{ranks}) \times 4(\text{gradient accumulation})$} \\
maximum gradient normalization & \multicolumn{4}{c}{0.1} \\
automatic mixed-precision & \multicolumn{4}{c}{BFloat16} \\
exponential moving average & \multicolumn{4}{c}{unused} \\
loss function & \multicolumn{4}{c}{MSE Loss} \\
\hline
Frames / Sampling Period & 64 / 2 & 64 / - & 64 / 2 & 64 / 2 \\
Input Size (L × C × H × W) & \multicolumn{4}{c}{64 × 1 × 112 × 112} \\
Patch Size ($\mathrm{p_s}$ / $\mathrm{p_t}$) & \multicolumn{4}{c}{16 / 1} \\
Enclosure Global Tokens & \multicolumn{4}{c}{$\mathrm{L_g=H_g=W_g}=3$} \\
Main Layers & \multicolumn{4}{c}{Encoder: (Mamba-3D-384) × 6 + Decoder: (Mamba-3D-384) × 2} \\
Masking Chain ($\mathrm{\gamma_s}$ / $\mathrm{\gamma_t}$) & 1 / 4 &  1 / 8 & 1 / 32 & 1 / 8 \\
Masking Ratio ($p_\mathrm{mask}$) & \multicolumn{4}{c}{0.8} \\
\bottomrule 
\end{tabular}
\label{tab:default_pretrain_configs}
\end{table*}


\begin{table*}[t]
\centering
\caption{\textbf{Default configurations and hyper-parameters of E-ViM³-p16 for downstream fine-tuning.}}
\belowrulesep=0pt
\aboverulesep=0pt
\small
\begin{tabular}{l|cccc}
\toprule
\hline
\multirow{2}{*}{Configuration \& Hyper-parameter} & \multicolumn{4}{c}{Datasets} \\
\cline{2-5}
& EchoNet-Dynamic & CAMUS & MICCAI-BUV & WHBUS \\
\midrule
optimizer & \multicolumn{4}{c}{AdamW~\cite{Loshchilov1} with $\beta_1=0.9, \beta_2=0.999, \varepsilon=1\mathrm{e}{-15}$} \\
weight decay & \multicolumn{4}{c}{$1\mathrm{e}{-1}$} \\
learning rate schedule & \multicolumn{4}{c}{cosine decay with linear warm-up} \\
basic learning rate & $1\mathrm{e}{-4}$ & $1\mathrm{e}{-4}$ & $5\mathrm{e}{-6}$ & $1\mathrm{e}{-5}$ \\
minimal learning rate & $1\mathrm{e}{-6}$ & $1\mathrm{e}{-6}$ & $5\mathrm{e}{-8}$ & $1\mathrm{e}{-7}$ \\
warm-up steps & 1 epoch & 1 epoch & 1 epoch & 1 epoch \\
total steps & 50 epochs & 100 epochs & 100 epochs & 100 epochs \\
total batch size & \multicolumn{4}{c}{$32=4(\text{batch size per rank}) \times 2(\text{ranks}) \times 4(\text{gradient accumulation})$} \\
maximum gradient normalization & \multicolumn{4}{c}{$1.0$} \\
automatic mixed-precision & \multicolumn{4}{c}{BFloat16} \\
exponential moving average & \multicolumn{4}{c}{every 5 steps after 100 steps with $\beta=0.9999$ and a warmup power of $0.75$} \\
loss function & L1 loss & L1 loss & weighted BCE Loss & weighted BCE Loss \\
\hline
Frames / Sampling Period & 64 / 2 & 64 / - & 64 / 2 & 64 / 2 \\
Input Size (L × C × H × W) & \multicolumn{4}{c}{64 × 1 × 112 × 112} \\
Patch Size ($\mathrm{p_s}$ / $\mathrm{p_t}$) & \multicolumn{4}{c}{16 / 1} \\
Enclosure Global Tokens & \multicolumn{4}{c}{$\mathrm{L_g=H_g=W_g}=3$} \\
Main Layers & \multicolumn{4}{c}{Backbone: (Mamba-3D-384) × 6 + Head: 3-layers-MLP} \\
Backbone output feature & \multicolumn{4}{c}{384 × 27 $\rightarrow$ 512} \\
Head MLP & \multicolumn{4}{c}{512 $\rightarrow$ 512 $\rightarrow$ 512 $\rightarrow$ 1} \\
\bottomrule
\end{tabular}
\label{tab:default_finetune_configs}
\end{table*}


\begin{table}[t]
\centering
\caption{\textbf{Results of the EF prediction task with 1\% training set labels on the EchoNet-Dynamic dataset.} “F / P” refers to input frames and sampling period. “\ddag” denotes a reproduction of hyper-parameters not used in the original method. The best and second-best results are \textbf{bolded} and \underline{underlined} respectively.}
\belowrulesep=0pt
\aboverulesep=0pt
\small
\begin{tabular}{l|c|ccc}
\toprule
Method & F / P & $\text{MAE}$$\downarrow$ & $\text{RMSE}$$\downarrow$ & $\text{R}^2$$\uparrow$ \\
\midrule
EchoNet~\cite{Ouyang2} & 32 / 2 & 8.56 & 11.83 & 0.06 \\
EchoCoTr~\cite{Muhtaseb1} & 36 / 4 & 8.18 & 11.08 & 0.18 \\
CoReEcho~\cite{Maani1} & 36 / 4 & 8.19 & 11.36 & 0.14 \\
\hline
EchoNet\ddag & 64 / 2 & 8.52 & 11.69 & 0.09 \\
EchoCoTr\ddag & 64 / 2 & 8.37 & 11.41 & 0.13 \\
CoReEcho\ddag & 64 / 2 & 8.28 & 11.29 & 0.15 \\
\hline
VideoMAE~\cite{Tong1} & 64 / 2 & 8.74 & 12.40 & 0.00 \\
MAE-ST~\cite{Feichtenhofer1} & 64 / 2 & 8.49 & 11.48 & 0.12 \\
AdaMAE~\cite{Bandara1} & 64 / 2 & 7.89 & 10.78 & 0.22 \\
VideoMamba~\cite{Li1} & 64 / 2 & 7.73 & 10.77 & 0.22 \\
Mamba-ND~\cite{Li2} w/ our SSP & 64 / 2 & 6.92 & 9.08 & 0.44 \\
\textbf{E-ViM³-p16 (Ours)} & 64 / 2 & \underline{6.13} & \underline{8.07} & \underline{0.56} \\
\textbf{E-ViM³-p4 (Ours)} & 64 / 2 & \textbf{6.01} & \textbf{7.98} & \textbf{0.57} \\
\bottomrule
\end{tabular}
\label{tab:s_echonet_part}
\end{table}


\begin{table}[t]
\centering
\caption{\textbf{Ablation study on the masking ratio on the EchoNet-Dynamic dataset.} The \colorbox{gray!20}{gray} row indicates the default setting for E-ViM³-p16. The best results are \textbf{bolded}.}
\belowrulesep=0pt
\aboverulesep=0pt
\small
\begin{tabular}{l|cc}
\toprule
Masking Ratio ($p_\mathrm{mask}$) & $\text{MAE}$$\downarrow$ & $\text{R}^2$$\uparrow$ \\
\midrule
0.5 & 4.0668 & 0.8057 \\
0.7 & 3.7884 & 0.8375 \\
\rowcolor{gray!20} 0.8 & \textbf{3.7035} & \textbf{0.8401} \\
0.9 & 3.9915 & 0.8093 \\
0.95 & 4.2157 & 0.7835 \\
\bottomrule
\end{tabular}
\label{tab:ablation_ratio}
\end{table}


\begin{table}[t]
\centering
\caption{\textbf{Ablation study on the input length and sampling strategy on the EchoNet-Dynamic dataset.} Use Enclosure Global Tokens as $\mathrm{L_g=H_g=W_g}=3$ unless otherwise specified. The \colorbox{gray!20}{gray} row indicates the default setting for E-ViM³-p16. The best results are \textbf{bolded}.}
\belowrulesep=0pt
\aboverulesep=0pt
\small
\begin{tabular}{ll|cc}
\toprule
Frames & Sampling Period & $\text{MAE}$$\downarrow$ & $\text{R}^2$$\uparrow$ \\
\midrule
16 & 2 & 4.6455 & 0.7330 \\
16 & 4 & 4.2055 & 0.7856 \\
32 & 2 & 3.8462 & 0.8247 \\
32 & 4 & 4.1343 & 0.7971 \\
64 & 1 & 4.0193 & 0.8076 \\
\rowcolor{gray!20} 64 & 2 & \textbf{3.7035} & \textbf{0.8401} \\
64 & 4 & 3.8573 & 0.8285 \\
128 & 1 & 3.8098 & 0.8356 \\
128 & 1 ($\mathrm{L_g=5}$) & 3.7904 & 0.8323 \\
128 & 2 & 3.8584 & 0.8267 \\
128 & 2 ($\mathrm{L_g=5}$) & 3.8234 & 0.8283 \\
\bottomrule
\end{tabular}
\label{tab:ablation_frames}
\end{table}


\begin{table}[t]
\centering
\caption{\textbf{Ablation study on the patch size ($\mathrm{p_s}$ / $\mathrm{p_t}$) and embedding dimension ($\mathrm{D}$) with different masking chain ($\mathrm{\gamma_s}$ / $\mathrm{\gamma_t}$) on the EchoNet-Dynamic dataset.} The \colorbox{gray!20}{gray} row indicates the default setting for E-ViM³-p16 and \colorbox{gray!50}{dark gray} for E-ViM³-p4. The best results are \textbf{bolded}.}
\belowrulesep=0pt
\aboverulesep=0pt
\small
\begin{tabular}{lll|cc}
\toprule
$\mathrm{p_s}$ / $\mathrm{p_t}$ & $\mathrm{D}$ & $\mathrm{\gamma_s}$ / $\mathrm{\gamma_t}$ & $\text{MAE}$$\downarrow$ & $\text{R}^2$$\uparrow$ \\
\midrule
16 / 1 & 96 & 1 / 2 & 4.1352 & 0.7923 \\
16 / 1 & 96 & 1 / 4 & 4.0149 & 0.8059 \\
16 / 1 & 96 & 2 / 2 & 4.1190 & 0.7958 \\
16 / 1 & 96 & 2 / 4 & 4.2361 & 0.7795 \\
16 / 1 & 192 & 1 / 4 & 3.9171 & 0.8227 \\
\rowcolor{gray!20} 16 / 1 & 384 & 1 / 4 & 3.7035 & 0.8401 \\
16 / 1 & 768 & 1 / 4 & 3.7550 & 0.8364 \\
\hline
4 / 2 & 96 & 1 / 4 & 3.7744 & 0.8356 \\
4 / 2 & 96 & 2 / 1 & 3.8007 & 0.8322 \\
4 / 2 & 96 & 2 / 2 & 3.7421 & 0.8411 \\
4 / 2 & 96 & 4 / 1 & 3.7808 & 0.8341 \\
4 / 2 & 96 & 4 / 2 & 3.7236 & 0.8387 \\
\rowcolor{gray!50} 4 / 2 & 192 & 4 / 2 & \textbf{3.5439} & \textbf{0.8536} \\
4 / 2 & 384 & 4 / 2 & 3.5974 & 0.8478 \\
\hline
16 / 2 & 384 & 1 / 2 & 3.8156 & 0.8329 \\
16 / 4 & 384 & 1 / 1 & 4.0516 & 0.8068 \\
4 / 1 & 192 & 4 / 4 & 3.6436 & 0.8495 \\
4 / 4 & 192 & 4 / 1 & 3.7823 & 0.8365 \\
\bottomrule
\end{tabular}
\label{tab:ablation_patch_size_dim}
\end{table}

\begin{table}[t]
\centering
\caption{\textbf{Ablation study on the number of encoder blocks ($\mathrm{N}$) of our E-ViM³-p16 on the EchoNet-Dynamic dataset.} “trainset keep ratio” is the percentage of trainset labels used for fine-tuning, corresponding to the main paper Table 4 (10\%) and the supplementary Table s4 (1\%). The \colorbox{gray!20}{gray} row indicates the default setting for E-ViM³-p16. The best results are \textbf{bolded}.}
\belowrulesep=0pt
\aboverulesep=0pt
\small
\begin{tabular}{cc|ccc}
\toprule
trainset keep ratio & $\mathrm{N}$ & $\text{MAE}$$\downarrow$ & $\text{RMSE}$$\downarrow$ & $\text{R}^2$$\uparrow$ \\
\midrule
 & 3 & 7.23 & 9.67 & 0.37 \\
1\% & 6 & 6.13 & 8.07 & 0.56 \\
 & 12 & 6.02 & 8.01 & 0.55 \\
\midrule
 & 3 & 4.54 & 6.07 & 0.75 \\
10\% & 6 & 4.17 & 5.57 & 0.79 \\
 & 12 & 4.15 & 5.51 & 0.80 \\
\midrule
 & 3 & 3.91 & 5.15 & 0.82 \\
\rowcolor{gray!20} 100\% & 6 & 3.70 & 4.89 & \textbf{0.84} \\
 & 12 & \textbf{3.68} & \textbf{4.82} & \textbf{0.84} \\
\bottomrule
\end{tabular}
\label{tab:ablation_encoder_block_number}
\end{table}

\clearpage


\clearpage
{
    \small
    \bibliographystyle{ieeenat_fullname}
    \bibliography{main}
}

\end{document}